\documentclass[runningheads]{llncs}
\usepackage[T1]{fontenc}
\usepackage{graphicx}

\usepackage{tikz}
\usepackage{comment}
\usepackage{amsmath,amssymb} 
\usepackage{color}

\usepackage{times}
\usepackage{soul}
\usepackage{url}
\usepackage[hidelinks]{hyperref}
\usepackage[utf8]{inputenc}
\usepackage[small]{caption}
\usepackage{subcaption}
\usepackage{booktabs}
\usepackage{algorithm}
\usepackage{algorithmic}
\usepackage{newfloat}
\usepackage{listings}
\usepackage{bm}
\usepackage{colortbl}
\usepackage{siunitx}
\usepackage{lipsum}
\usepackage{multirow}
\usepackage{tabularx}
\usepackage{adjustbox}
\usepackage{xspace}
\usepackage{bbm}
\usetikzlibrary{bayesnet}
\usepackage{wrapfig}
\usepackage{subcaption}

\newcolumntype{L}{>{\raggedright\arraybackslash}X}
\floatstyle{ruled}
\newfloat{listing}{tb}{lst}{}
\floatname{listing}{Listing}

\newcommand{\model}{\texttt{SAViR-T}\xspace}

\renewcommand*{\figureautorefname}{Fig.}
\renewcommand*{\tableautorefname}{Tab.}
\renewcommand*{\sectionautorefname}{Sec.}

\renewcommand*{\equationautorefname}{Eq.}

\begin{document}
\title{SAViR-T: Spatially Attentive Visual Reasoning with Transformers}
%
%
\vspace{-1.5em}
\author{\normalsize{Pritish Sahu}  \and \normalsize{Kalliopi Basioti}  \and \normalsize{Vladimir Pavlovic} }
\institute{\normalsize{Dept. of Computer Science, Rutgers University, NJ, USA},  \\
\email{\normalsize{\texttt{pritish.sahu@rutgers.edu}, \texttt{kib21@scarletmail.rutgers.edu}, \texttt{vladimir@cs.rutgers.edu} }
}}
\vspace{-1.5em}

%
%
\maketitle              

\definecolor{redcol}{rgb}{1, 0, 0}
\definecolor{bluecol}{rgb}{0, 0, 1}
\newcommand{\red}[1]{\textcolor{redcol}{#1}}
\newcommand{\blue}[1]{\textcolor{bluecol}{#1}}
\renewcommand{\paragraph}[1]{\smallskip\noindent{\bf{#1}}}

\newcommand{\todo}[1]{\red{TODO: {#1}}}
\newcommand{\colons}[1]{``{#1}''}
\newcommand{\tb}[1]{\textbf{#1}}
\newcommand{\mb}[1]{\mathbf{#1}}
\newcommand{\bs}[1]{\boldsymbol{#1}}
\def\ith#1{#1^\textit{th}}

\renewcommand{\vec}[1]{\mathbold{#1}}
\newcommand{\mat}[1]{\mathbold{#1}}
\newcommand{\vx}{\vec{x}}
\newcommand{\vX}{\mat{X}}

\newcommand{\secref}[1]{Section~\ref{sec:#1}}
\newcommand{\figref}[1]{Figure~\ref{fig:#1}}
\newcommand{\tabref}[1]{Table~\ref{tab:#1}}
\newcommand{\eqnref}[1]{\eqref{eq:#1}}
\newcommand\RotText[1]{\rotatebox[origin=c]{90}{\parbox{1cm}{\centering#1}}}

\def\algorithmautorefname{Algorithm}
\def\figureautorefname{Figure}
\def\tableautorefname{Table}
\def\equationautorefname{Eq.}
\def\sectionautorefname{Section}
\def\etal{et~al.\_} 
\def\eg{e.g.,~} 
\def\ie{i.e.,~} 
\def\etc{etc} 
\def\cf{cf.~} 
\def\viz{viz.~} 
\def\vs{vs.~} 
\def\newtext#1{\textcolor{blue}{#1}}
\def\modtext#1{\textcolor{red}{#1}}

\begin{abstract}
We present a novel computational model, \texttt{SAViR-T}, for the family of visual reasoning problems embodied in the Raven's Progressive Matrices (RPM). 
Our model considers explicit spatial semantics of visual elements within each image in the puzzle, encoded as spatio-visual tokens, and learns the intra-image as well as the inter-image token dependencies, highly relevant for the visual reasoning task. Token-wise relationship, modeled through a transformer-based \model architecture, extract group (row or column) driven representations by leveraging the group-rule coherence and use this as the inductive bias to extract the underlying rule representations in the top two row (or column) per token 
in the RPM. We use this relation representations to locate the correct choice image that completes the last row or column for the RPM. Extensive experiments across both synthetic RPM benchmarks, including RAVEN, I-RAVEN, RAVEN-FAIR, and PGM, and the natural image-based "V-PROM" demonstrate that \model sets a new state-of-the-art for visual reasoning, exceeding prior models' performance by a considerable margin. 

\end{abstract}

\section{Introduction}


\begin{figure}[!ht]
    \begin{center}
        \includegraphics[width=0.7\linewidth]{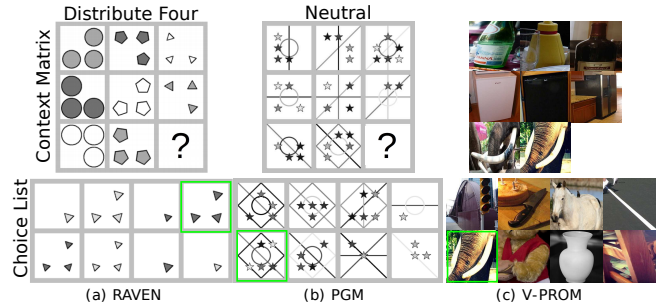}
    \end{center}
    \caption{Three RPM examples from the RAVEN (a), PGM (b), and V-PROM (c) datasets. The highlighted green box in the choice list represents the correct image to be placed at the missing location of the context matrix. Solving these RPM requires identifying the underlying rules applied to image attributes or lines along a row or column and,  the image with the best fit from the choice list to the rules is the correct answer. For example, the rules governing RAVEN are ``distribute\_three" on $\{$position, color $\}$ and ``progression" on $\{$type, size$\}$. The rules governing PGM instances are ``OR" on object type line and ``XOR" on the position of shapes. And the row rule governing the V-PROM example is "And". }
    \label{fig:rpm}
\end{figure}

Human abstract reasoning is the analytic process aimed at decision-making or solving a problem~\cite{apps2008abstract}. In the realm of visual reasoning, humans find it advantageous, explicitly or implicitly, to break down an image into well-understood low-level concepts before proceeding with the reasoning task, e.g., examining object properties or counting objects. These low-level concepts are combined to form high-level abstract concepts in a list of images, enabling relational reasoning functions such as assessing the increment of the object count, changes in the type of object, or object properties, and subsequently applying the acquired knowledge to unseen scenarios. However, replicating such reasoning processes in machines is particularly challenging~\cite{lake2017building}. 

A popular test format of abstract reasoning in visual domain is the Raven's Progressive Matrix (RPM), developed on Spearman's work on human general intelligence~\cite{raven1998raven}. The test is designed as an incomplete $3\times 3$ matrix, with each matrix element being an image and the bottom right location left empty, c.f., \autoref{fig:rpm}. 
Every image can contain one or more objects or lines characterized by the attributes  shape, color, scale, rotation angle, and, holistically, the counts of items or their variability (all circles, all pentagons, both pentagons and circles, etc.). The top two rows or columns follow a certain unknown rule applied to the attributes; the task is to pick the correct image from an unordered set of choices, satisfying the same constraints. For example, in \autoref{fig:rpm} we present three instances of RPM, RAVEN~\cite{zhang2019raven}, PGM~\cite{barrett2018measuring}, and V-PROM~\cite{teney2020v} datasets, where first eight images are denoted as context images, and below them is the set of choice images. 

Classical computational models for solving RPMs, built upon access to symbolic attribute representations of the images~\cite{carpenter1990one,lovett2007analogy,lovett2010structure,lovett2009solving}, are incapable of adapting to unseen domains. The success of deep models in other computer vision tasks made it possible to exploit the feature representation, and relational learning concepts in visual reasoning~\cite{malkinski2022review}. Initial studies~\cite{zhang2019raven,barrett2018measuring} using widely popular neural network architecture such as ResNet~\cite{he2016deep} and LSTM~\cite{hochreiter1997long} failed in solving general reasoning tasks. These models aim to discover underlying rules by directly mapping the eight context images to each choice image. 
Modeling the reasoning network~\cite{hu2020stratified,zhang2019learning,benny2021scale} similar to the reasoning process by human beings has shown huge performance improvement. All recent works, utilize an encoding mechanism to extract the features/attributes of single or groups of images, followed by a reasoning model that learns the underlying rule from the extracted features to predict scores for images in choice list. This  to contrast against each choice image and elucidate the best image in the missing location. However, these models make use of holistic image representations, which ignore the important local, spatially contextualized features. 
Because typical reasoning patterns make use of intra and inter-image object-level relationships, holistic representations are likely to lead to suboptimal reasoning performance of the models that rely on it.



In this work, we focus on using local, spatially-contextualized features accompanied with an attention mechanism to learn the rule constraint within and across groups (rows or columns). We use a bottom-up and top-down approach to the visual encoding and the reasoning process. From bottom-up, we address how a set of image regions are associated with each other via the self-attention mechanism from visual transformers. Specifically, instead of extracting a traditional holistic feature vector on image-level
, we constrain semantic visual tokens to attend to different image patches. The top-down process is driven to solve visual reasoning tasks that predict an attention distribution over the image regions. To this end, we propose ``\model",  Spatially Attentive Visual Reasoning with Transformers, that naturally integrates the attended region vectors with abstract reasoning. Our reasoning network focuses on entities of interest obtained from the attended vector, since the irrelevant local areas have been filtered out. Next, the reasoning task determines the Principal Shared rules in the two complete groups (typically, the top two rows) of the RPM, per local region, which are then fused to provide an integrated rule representation. We define a similarity metric to compare the extracted rule representation with the rules formed in the last row when placing each choice at the missing location. The choice with the highest score is predicted as the correct answer. 
Our contributions in this work are three-fold:
\begin{itemize}
    \item We propose a novel abstract visual reasoning model, \model, using spatially-localized attended features, for reasoning tasks. \model accomplishes this using the Backbone Network, responsible for extracting a set of image region encodings, the Visual Transformer, performing self-attention on the tokenized feature maps, and the Reasoning Network, which elucidates the rules governing the puzzle, over row-column groups, to predicting the solution to the RPM.
    
    \item \model automatically learns to focus on different semantic regions of the input images, addressing the problem of extracting holistic feature vectors per image, which may omit critical objects at finer visual scales. Our approach is generic because it is suitable for any configuration of the RPM problems without the need to modify the model for different image structures.

    \item We drastically improved the reasoning accuracy overall RPM benchmarks, echoed in substantial enhancement on the ``$3\times 3$ Grid", ``Out Single, In Single," and ``Out Single, In Four Distribute" configurations for RAVEN and I-RAVEN, with strong accuracy gains in the other configurations. We show an average improvement of $2-3 \%$ for RAVEN-type and PGM datasets. Performance improvement of \model on V-PROM, a natural image RPM benchmark, significantly improves by $10 \%$  over the current state-of-the-art models.
\end{itemize}
\section{Related Works}
\subsection{Abstract Visual Reasoning}
RPM is a form of non-verbal assessment for human intelligence with strong roots in cognitive science~\cite{carpenter1990one,barrett2018measuring,zhang2019raven}.
It measures an individual's eductive ability, i.e.,  the ability to find patterns in the apparent chaos of a set of visual scenes~\cite{raven1998raven}. RPM consists of a context matrix of size $3\times 3$ that contains eight images with a missing image at the last row last column, and the participant has to locate the correct image from a choice set of size eight. In the early stages, RPM datasets were created manually, and the popular computational models solved them using hand-crafted feature representations~\cite{mcgreggor2014confident}, or access to symbolic representations~\cite{lovett2007analogy}. It motivated the need for large-scale RPM datasets and the requirement for efficient reasoning models that utilized minimal prior knowledge.
The first automatic RPM generation \cite{wang2015automatic} work was based on using first-order logic, followed by two large-scale RPM datasets RAVEN~\cite{zhang2019raven} and Procedurally Generated Matrices (PGM)~\cite{barrett2018measuring}. However, the RAVEN dataset contained a hidden shortcut solution where a model trained on the choice set only can achieve better performance than many state-of-the-art models. The reason behind this behavior is rooted in the creation of the choice set. Given the correct image, the distractor images were derived by randomly changing  only one attribute. 
In response, two modified versions of the dataset, I-RAVEN by SRAN~\cite{hu2020stratified} and RAVEN-FAIR~\cite{benny2021scale}, were proposed to remove the shortcut solution and increase the difficulty levels of the distractors. Both the works, devised algorithm to generate a different set of distractors and provide evidence through experiments to claim the non-existence of any shortcut solution. 
The first significant advancement in RPM was by Wild Relational Network (WReN)~\cite{barrett2018measuring}, which applies the relation network of~\cite{santoro2017simple} multiple times to solve the abstract reasoning problem. LEN~\cite{zheng2019abstract} learns to reason using a triplet of images in a row or column as input to a variant of the relation network.
This work empirically supports improvements in performance using curriculum and reinforcement learning frameworks. CoPINet~\cite{zhang2019learning} suggests a contrastive learning algorithm to learn the underlying rules from given images. SRAN~\cite{hu2020stratified} designs a hierarchical rule-aware neural network framework that learns rule embeddings through a series of steps of learning image representation, followed by row representation, and finally learning rules by pairing rows. 

\subsection{Transformer in Vision}
Transformers~\cite{vaswani2017attention} for machine translation have become widely adopted in numerous NLP tasks~\cite{devlin2018bert,radford2018improving,brown2020language,liu2019roberta}. A transformer consists of self-attention layers added along with MLP layers. The self-attention mechanism plays a key role in drawing out the global dependencies between input and output. Grouped with the non-sequential processing of sentences, transformers demonstrate superiority in large-scale training scenarios compared to Recurrent Neural Networks. It avoids a drop in performance due to long-term dependencies. Recently, there has been a steady influx of visual transformer models to various vision tasks: image classification~\cite{chen2020generative,dosovitskiy2020image}, object detection~\cite{carion2020end,dai2021up,sun2021rethinking}, segmentation~\cite{wang2021end}, image generation~\cite{parmar2018image}, video processing~\cite{zeng2020learning,zhou2018end}, VQA~\cite{li2019visualbert}. Among them, Vision Transformer (ViT)~\cite{dosovitskiy2020image} designed for image classification has closed the gap with the performance provided by any state-of-the-art models (e.g., ImageNet, ResNet) based on convolution. Similar to the word sequence referred to as tokens required by Transformers in NLP, ViT splits an image into patches and uses the linear embeddings of these patches as an input sequence. Our idea is closely related to learning intra-sequence relationships that can be captured automatically through self-attention.

\section{Method}
Before presenting our reasoning model, in \autoref{sec:rpm_desc} we provide a formal description of the RPM task designed to measure abstract reasoning. The description articulates the condition required in a reasoning model to solve RPM questions successfully. In \autoref{sec:model}, we describe the three major components of our \model: the backbone network, the visual transformer, and the reasoning network. Our objective is based on using local feature maps as tokens for rule discovery among visual attributes along rows or columns to solve RPM questions.

\subsection{Raven’s Progressive Matrices}\label{sec:rpm_desc}
\begin{figure}[!ht]
    \begin{center}
        \includegraphics[width=\linewidth]{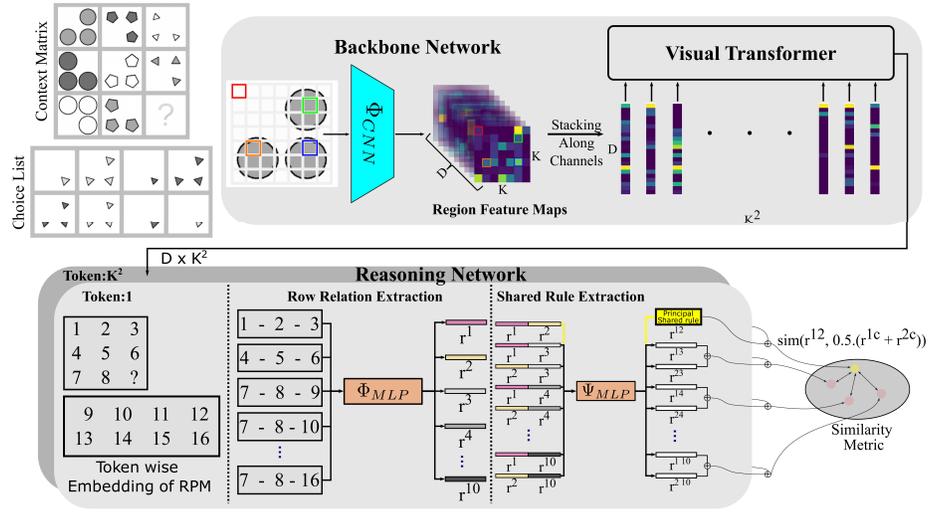}
    \end{center}
    \caption{
    \model consists of \textbf{Backbone Network}, \textbf{Visual Transformer}, \textbf{Reasoning Network}. Each image in $\mathcal{M}$ given to the Backbone Network $\Phi_{\text{CNN}}$ to extract the ``Feature Maps", $f \in \mathbb{R}^{D \times K \times K}$. Visual Transformer attends on the features of local image patches and returns the attended vectors $\hat{f}$ of each image. The Reasoning Network functions per patch (depicted as $K$ parallel layers) for the entire context and choice attended vectors. Per patch, we start with group (row or column) rule extraction $r^i$ via $\Phi_{\text{MLP}}$, followed by shared     Shared rule extraction $r^{ij}$ via $\Psi_{\text{MLP}}$. The Principal Shared rule $r^{12}$ is compared against extracted Shared rule for choice $a$, $\frac{1}{2}(r^{1a}+r^{2a})$. 
    The choice image with max similarity score is predicted as the answer.
    }
    \label{fig:model}
\end{figure}

Given a list of observed images in the form of Raven's Progressive Matrix ($\mathcal{M}$)  referred to as the context of size $3 \times 3$ with a missing final element at $\mathcal{M}_{3,3}$, where $\mathcal{M}_{i,j}$ denotes the $j$-th image at $i$-th row. In \autoref{fig:model}, for ease of notation we refer to the images in $\mathcal{M}$ as index location 1 through 16 as formatted in the dataset, where the first eight form the context matrix and the rest belong to the choice list. The task of a learner is to solve the context $\mathcal{M}$ by finding the best-fit answer image from an unordered set of choices $\mathcal{A} = \{a_1, \ldots, a_8\}$. 
The images in an RPM can be decomposed into attributes, objects, and the object count. Learning intra-relationship between these visual components will guide the model to form a stronger inter-relationship between images constrained by rules.
The learner needs to locate objects in each image, extract their visual attributes such as color, size, and shape,  lastly infer the rules ``$r$" such as ``constant," ``progression," ``OR," etc., that satisfies the attributes among a list of images. Usually, these rules ``$r$" are applied either row-wise or column-wise according to ~\cite{carpenter1990one} on the decomposed visual elements mentioned above. Since an RPM is based on a set of rules applied either row-wise or column-wise, the learner needs to pick the shared rules between the top two rows or columns. Among the choice list, the correct image from $\mathcal{A}$ that, when placed at the missing location in the last row or column, will satisfy these shared rules.

\subsection{Our Approach: \model}\label{sec:model}
Our method consists of three sub-modules: (i) a Backbone Network,  (ii) a Visual Transformer (VT), and (iii) a Reasoning Network, trained end to end. Please refer \autoref{fig:model} for an illustration of our model. First, we process the images in $\mathcal{M}$ via several convolutional blocks referred to as the backbone network.  The output feature map is given as to the visual transformer to extract the attended visual embedding. The attended embedding is given to the reasoning network to discover the embedded rule representation in RPM. A scoring function is used to rank the choice images by comparing its row or column representation (extracted by placing it in the missing location) with the rule representation and predicting the index with the highest similarity as the correct choice. We leverage the strength of convolutions, which learns location invariant low-level neighborhood structures and visual transformer to relate to the higher-order semantic concepts. We treat each local region as a token separately in reasoning and applying a fusion function to recover the hidden rules that point the model to the correct answer. 

\subsubsection{Backbone Network} 
The backbone network receives as input an image from the context ($\mathcal{M}$) or choice list ($\mathcal{A}$) of size $\mathbb{R}^{C \times H \times W }$. The extracted feature map ($f_{ij},f_a$) is of dimension $\mathbb{R}^{D \times K \times K}$ where $K \times K$  is the number of image regions also referred to as tokens, and $D$ is the dimension of the feature vector of each area. Accordingly, each feature vector corresponds to a $(\frac{W}{K} \times \frac{H}{K})$ pixel region retaining the spatial information of the raw image. We intend to summarize the high-level semantic information present in the image by learning from a set of low-level visual tokens. To this effect, we employ several convolutional blocks as our backbone network,  denoted as $\Phi_{\text{CNN}}$, to extract local features from images.
We use ResNet~\cite{he2016deep} as our primary backbone network, although we show results with other popular backbones in our ablation study. 
\begin{align}
f_{ij} &= \Phi_{\text{CNN}}(\mathcal{M}_{ij}), \quad i,j\in \{1,3\}, i,j\neq3 \\
f_{a} &=  \Phi_{\text{CNN}}(a),   \quad \forall a \in \mathcal{A}
\end{align}
Both the context and choice feature representation are generated using the same  network. We flatten and concatenate the matrix format in context to prepare feature vector $F_{\mathcal{M}} = [f_{11},\ldots,f_{32}] \in \mathbb{R}^{8\times K^2 \times D}$, where each feature map is reshaped into a $K^2$-tall sequence of tokens, $\mathbb{R}^{K^2 \times D}$. Choices are processed in the same manner, $F_{\mathcal{A}} = [f_{a_1},\ldots,f_{a_8}] \in \mathbb{R}^{8\times K^2 \times D}$. Finally, both the context and the choice representations are concatenated as $F = [F_{\mathcal{M}}, F_{\mathcal{A}}]$, with $[\quad]$ denoting the concatenation operator.
\subsubsection{Visual Transformer} 
To learn the concepts responsible for reasoning, we seek to model the interactions between local regions of an image as a bottom-up process, followed by top-down attention to encoder relations over the regions.  We adopt Visual Transformer ~\cite{dosovitskiy2020image}, which learns the attention weights between tokens to focus on relationally-relevant regions within images.
The  transformer is composed of ``Multi-head Self-Attention" (MSA) mechanism followed by a multi-layer perceptron (MLP). Both are combined together in layers $l=1,\ldots,L$ to form the transformer encoder. 
A layer normalization layer (LN) and residual connection are added before and after every core component. The interactions between the tokens generate an attention map for each layer and head. Below we describe the steps involved in learning the attended vectors. 
\subsubsection{Reasoning Model} 
Human representations of space are believed to be hierarchical \cite{palmer1977hierarchical,palmer1994rethinking}, with objects parsed into parts and grouped into part constellations. 
To mimic this, \model generates weighted representations over local regions in the image as described above. Our reasoning module combines the inductive bias in an RPM and per-patch representations to learn spatial relations between images. We start by translating these attended region vectors obtained from RPM above into within-group relational reasoning instructions, expressed in terms of the row representations $r_k^i$ via $\Phi_{\text{MLP}}$. These representations hold knowledge of the rules that bind the images in the $i$-th row. We realize ten row representations, including eight possible last rows, where each choice is replaced at the missing location (similarly for column). We define function $\Psi_{\text{MLP}}$ that retrieves the common across-group rule representations $r_k^{ij}$, given the pair $(r_k^i,r_k^j)$. The maximum similarity score between the last-row rules based on Choice list, $\{r_k^{1c},r_k^{2c}\}, c=3,\ldots,10$, and the extracted Principal Shared rule from the top two rows $r_k^{12}$ indicates the correct answer. 

\textbf{Row Relation Extraction Module.} Since our image encoding is prepared region-wise, we focus on these regions independently to detect patterns maintaining the order of RPM, i.e., row-wise (left to right) or column-wise (top to bottom).  We restructure the resulting output of a RPM from the transformer $\hat{F}$ as $\mathbb{R}^{K^2 \times 16 \times D}$,  where the emdeddings of each local region $k$ for an image at index $n$ in the RPM is denoted $\hat{f}^n_{k}$ for $n = \{1, 2, \ldots, 16\}$\footnote{$16 = 8+ 8$ for eight Context and eight Choice images of an RPM.} and $k = \{1, \ldots, K^2\}$. We collect theses embeddings for every row and column as triplets ($\hat{f}^{1}_{k}, \hat{f}^{2}_{k}, \hat{f}^{3}_{k}$), ($\hat{f}^{4}_{k}, \hat{f}^{5}_{k}, \hat{f}^{6}_{k}$) , ($\hat{f}^{7}_{k}, \hat{f}^{8}_{k}, \hat{f}^{a}_{k}$), $a =9,\ldots,16$. Similarly, the triplets formed from columns are ($\hat{f}^{1}_{k}, \hat{f}^{4}_{k}, \hat{f}^{7}_{k}$), ($\hat{f}^{2}_{k}, \hat{f}^{5}_{k}, \hat{f}^{8}_{k}$) , ($\hat{f}^{3}_{k}, \hat{f}^{6}_{k}, \hat{f}^{a}_{k}$). Each row and column is concatenated along the feature dimension and processed through a relation extraction function $\Phi_{MLP}$:
\begin{equation}
\begin{aligned}
    r^1_{k} &= \Phi_{\text{MLP}}(\hat{f}^{1}_{k}, \hat{f}^{2}_{k}, \hat{f}^{3}_{k}), \quad c^1_{k} = \Phi_{\text{MLP}}(\hat{f}^{1}_{k}, \hat{f}^{4}_{k}, \hat{f}^{7}_{k}) \\
    r^2_{k} &= \Phi_{\text{MLP}}(\hat{f}^{4}_{k}, \hat{f}^{5}_{k}, \hat{f}^{6}_{k}), \quad c^2_{k} = \Phi_{\text{MLP}}(\hat{f}^{2}_{k}, \hat{f}^{5}_{k}, \hat{f}^{8}_{k}) \\
    r^3_{k} &= \Phi_{\text{MLP}}(\hat{f}^{7}_{k}, \hat{f}^{8}_{k}, \hat{f}^{a}_{k}), \quad c^3_{k} = \Phi_{\text{MLP}}(\hat{f}^{3}_{k}, \hat{f}^{6}_{k}, \hat{f}^{a}_{k}).
\end{aligned}    
\end{equation}
The function $\Phi_{\text{MLP}}$ has two-fold aims: (a) it seeks to capture common properties of relational reasoning along a row/column for each local region; (b) it updates the attention weights to cast-aside the irrelevant portions that do not contribute to rules.  

\textbf{Shared Rule Extraction Module.} 
The common set of rules conditioned on visual attributes in the first two rows of a RPM is the Principal Shared rule set which the last row has to match to select the correct image in the missing location. The top two rows will contain these common sets of rules, possibly along with the rules unique to their own rows. Given $r^i_k$, $r^j_k$, the goal of function $\Psi_{MLP}$ is to extract these Shared rules between the pairs of rows/columns, 
\begin{align} \label{eq:rc}
    r^{ij}_k = \Psi_{\text{MLP}}(r^i_k, r^j_k), \quad c^{ij}_k = \Psi_{\text{MLP}}(c^i_k, c^j_k), \quad rc_k^{ij} = [ r_k^{ij}, c_k^{ij} ].
\end{align}
Similar to the idea behind $\Phi_{\text{MLP}}$, function $\Psi_{\text{MLP}}$ aims to elucidate the Shared relationships between any pair of rows or columns $i, j$. Those relationships should hold across image patches, thus we fuse $rc_k^{ij}$ obtained for $k=1,\ldots,K^2$ regions using averaging, leading to the Shared rule embedding $rc^{ij} = \frac{1}{K^2}\sum_k rc_k^{ij}$ over the entire image.  

\subsection{Training and Inference}
Given the principal Shared rule embedding $rc^{1,2}$ from the top two row pairs, a similarity metric is a function of closeness between $\text{sim}(rc^{1,2}, rc^{a}))$, where $rc^{a} = \frac{1}{2}(rc^{1a} + rc^{2a})$ is the average of the Shared rule embeddings among the choice $a$ and the top two rows, and $\text{sim}(\cdot,\cdot)$ is the inner product between $rc^{12}$ and the average Shared rule embedding.
The similarity score will be higher when the correct image is placed at the last row compared to the wrong other choice,
\begin{equation}\label{eq:best_answer_inference}
a^* = \arg \max_{a}  \text{sim}(rc^{12}, rc^{a}).
\end{equation}
We use cross entropy as our loss function to train \model end-to-end. To bolster generalization property of our model, two types of augmentation were adapted from~\cite{zhang2019learning}: (i) shuffle the order of top two rows or columns, as the resulting change will not affect the final solution since the rules remain unaffected; and (ii) shuffling the index location of correct image in the unordered set of choice list.
After training our model, we can use \model to solve new RPM problems (i.e., during testing) by applying \eqref{eq:best_answer_inference}.


\section{Experiments}
We study the effectiveness of our proposed \model for solving the challenging RPM questions, specifically focusing on PGM~\cite{barrett2018measuring}, RAVEN~\cite{zhang2019raven}, I-RAVEN~\cite{hu2020stratified}, RAVEN-FAIR~\cite{benny2021scale}, and V-PROM~\cite{teney2020v}. Details about the datasets can be found in the Supplementary. Next, we describe the experimental details of our simulations, followed by the results of these experiments and the performance analysis of obtained results, including an ablation study of our \model.

\subsection{Experimental Settings}
\label{sec:datasets}
\textbf{Experimental Settings}. We trained our model for 100 epochs on all three RAVEN datasets and for 50 epochs for PGM, where each RPM input sample is scaled to $16 \times 224 \times 224$. We use the validation set to measure model performance during the training process and report the accuracy on the test set for the checkpoint with the best validation accuracy. We keep the architectural hyperparameters same for all models across all the tasks. For \model, we adopt ResNet-18 as our backbone for results in \tabref{Raven_IRaven} and \tabref{pgm}. We set depth to one and counts of heads to 3 for the transformer for RAVEN datasets and heads to 6 for PGM. Finally, in our reasoning module, we use a two-layer MLP, $\Phi_{\text{MLP}}$ and a four-layer MLP, $\Psi_{\text{MLP}}$ with a dropout of 0.5 applied to the last layer. As the rules are applied row-wise in all RAVEN datasets, we set the column vector $c^{ij}_k$ in \eqref{eq:rc} as zero-vector while training our model. No changes are made while training for PGM, as the tuple (rule, object, attrbute) can be applied either along the rows or columns. All our modules are implemented in PyTorch and optimized using ADAM optimizer, with the learning rate of $10^{-4}$ and decay parameters set as $\beta_1= 0.9$, $\beta_2 = 0.999$, and $\epsilon = 10^{-8}$. 

For V-PROM dataset we train \model and baseline models for 100 epochs. As the authors proposed in \cite{teney2020v} instead of using the raw images $224\times 224\times 3$, we use the features extracted from the PyTorch pretrained ResNet-101 before the last average pooling layer; this means that we use for each image a $2048\times 7 \times 7$ representation, which translates to 49 tokens in \model. To further reduce the complexity of our model, we use an MLP layer to derive $512\times7\times 7$ feature vectors; we must mention that this MLP becomes part of \model where we learn its parameters during training. Also, in the Transformer we use five heads. For the remaining parts we follow the same architecture as for RAVEN-based datasets.


\setlength{\tabcolsep}{4pt}
\begin{table}[t!]
\vspace{-1.6em}
\begin{center}
\caption{\hspace*{0mm}Model performance (\%) on RAVEN / I-RAVEN. 
}

\resizebox{\columnwidth}{!}{
\begin{tabular}{l}
\begin{tabular}{ccccccccc}
    \hline\noalign{\smallskip}
    Method & Acc & \texttt{Center} &  \texttt{2$\times$2Grid}  & \texttt{3$\times$3Grid}  & \texttt{L-R}& \texttt{U-D} & \texttt{O-IC} & \texttt{O-IG} \\
    \noalign{\smallskip}
    \hline
    \noalign{\smallskip}
    LSTM~\cite{hochreiter1997long} &  13.1/18.9 & 13.2/26.2 & 14.1/16.7 & 13.7/15.1 & 12.8/14.6 & 12.4/16.5 & 12.2/21.9 & 13/21.1 \\
    \noalign{\smallskip}
  WReN~\cite{barrett2018measuring} & 34.0/23.8 & 58.4/29.4 & 38.9/26.8 & 37.7/23.5 & 21.6/21.9 & 19.7/21.4 & 38.8/22.5 & 22.6/21.5 \\
    \noalign{\smallskip}
ResNet+DRT~\cite{zhang2019raven} & 59.6/40.4 & 58.1/46.5 & 46.5/28.8 & 50.4/27.3 & 65.8/50.1 & 67.1/49.8 & 69.1/46.0 & 60.1/34.2 \\
    \noalign{\smallskip}
    LEN~\cite{zheng2019abstract} & 72.9/39.0 & 80.2/45.5 & 57.5/27.9 & 62.1/26.6 & 73.5/44.2 & 81.2/43.6 & 84.4/50.5 & 71.5/34.9 \\
    \noalign{\smallskip}
CoPINet~\cite{zhang2019learning} & 91.4/46.1 & 95.1/54.4 & 77.5/36.8 & 78.9/31.9 & 99.1/51.9 & 99.7/52.5 & 98.5/52.2 & 91.4/42.8 \\
    \noalign{\smallskip}
    SRAN~\cite{hu2020stratified} & 56.1$^\dag$/60.8 & 78.2$^\dag$/78.2 & 44.0$^\dag$/50.1 & 44.1$^\dag$/42.4 & 65.0$^\dag$/70.1 & 61.0$^\dag$/70.3 & 60.2$^\dag$/68.2 & 40.1$^\dag$/46.3 \\
    \noalign{\smallskip}
DCNet~\cite{zhuo2020effective} & 93.6/49.4 & 97.8/57.8 & 81.7/34.1 & \textbf{86.7}/35.5 & \textbf{99.8}/58.5 & \textbf{99.8}/60 & \textbf{99.0}/57.0 & \textbf{91.5}/42.9 \\
   \noalign{\smallskip}
SCL~\cite{wu2020scattering} &  91.6/95.0 & \textbf{98.1}/99.0 & 91.0/96.2 & 82.5/89.5 & 96.8/97.9 &  96.5/97.1 & 96.0/97.6 & 80.1/87.7 \\
   \noalign{\smallskip}

    \model (Ours) & \textbf{94.0}/\textbf{98.1} & 97.8/\textbf{99.5} & 94.7/\textbf{98.1} & 83.8/\textbf{93.8} & 97.8/\textbf{99.6} & 98.2/\textbf{99.1} & 97.6/\textbf{99.5} & 88.0/\textbf{97.2} \\
    \noalign{\smallskip}
    \hline
    \noalign{\smallskip}
    Human & 84.41/- & 95.45/- & 81.82/- & 79.55/- & 86.36/- & 81.81/- & 86.36/- & 81.81/- \\
    \noalign{\smallskip}
    \hline
\end{tabular}\\
\rule{0in}{1.2em}$^\dag$\scriptsize indicates our evaluation of the baseline in the absence of published results.\\
\end{tabular}
}
\label{tab:Raven_IRaven}
\hfill
\end{center}
\end{table}
\setlength{\tabcolsep}{3.0pt}
\begin{table}[t!]
\vspace{-1.5em}
\begin{center}
\caption{\hspace*{0mm}Test accuracy of different models on PGM.}
\resizebox{\columnwidth}{!}{%
\begin{tabular}{ccccccccccc}
    \hline\noalign{\smallskip}
    Method & LSTM~\cite{hochreiter1997long} & ResNet  & CoPINet~\cite{zhang2019learning}  & WReN~\cite{barrett2018measuring} & MXGNet~\cite{wang2020abstract} & LEN~\cite{zheng2019abstract} & SRAN~\cite{hu2020stratified} & DCNet~\cite{zhuo2020effective} & SCL~\cite{wu2020scattering} & \model (Ours) \\
    \noalign{\smallskip}
    \hline
    \noalign{\smallskip}
    Acc & 35.8 & 42.0 & 56.4 & 62.6 & 66.7 & 68.1 & 71.3 & 68.57 & 88.9 & \textbf{91.2} \\
    \noalign{\smallskip}
    \hline
\end{tabular}%
}
\label{tab:pgm}
\end{center}
\hfill
\vspace{-1.0em}
\end{table}
\setlength{\tabcolsep}{1.4pt}
\setlength{\tabcolsep}{20pt}
\begin{table}[t!]
\begin{center}
\caption{\hspace*{0mm}Test accuracy of different models on V-PROM.
}
\resizebox{0.98\columnwidth}{!}{
\begin{tabular}{l}
\begin{tabular}{ccccccc}
    \hline\noalign{\smallskip}
    Method  & RN~\cite{teney2020v} & DCNet~\cite{zhuo2020effective} & SRAN~\cite{hu2020stratified} & \model (Ours) \\ 
    \noalign{\smallskip}
    \hline
    \noalign{\smallskip}
    Acc & 52.83${}^\dag{}^\dag$ & 30.39$^\dag$ & 40.84$^\dag$ & \textbf{62.62}   \\
    \noalign{\smallskip}
    \hline
\end{tabular}%
\\
\rule{0in}{1.2em}$^\dag$\scriptsize indicates our evaluation of the baseline in the absence of published results.\\ \rule{0in}{1.2em}${}^\dag{}^\dag$\scriptsize indicates our evaluation of the baseline (reported result in \cite{teney2020v} was 51.2).
\end{tabular}
}
\label{tab:vprom}
\end{center}
\hfill
\vspace{-2.5em}
\end{table}
\setlength{\tabcolsep}{1.4pt}
\setlength{\tabcolsep}{10pt}
\begin{table}
\vspace{-1.5em}
\begin{center}
\caption{\hspace*{0mm}Test accuracy of different models on RAVEN-FAIR.}
\resizebox{0.98\columnwidth}{!}{
\begin{tabular}{l}
\begin{tabular}{ccccccc}
    \hline\noalign{\smallskip}
    Method  & ResNet~\cite{he2016deep} & LEN~\cite{zheng2019abstract} & COPINet~\cite{zhang2019learning} & DCNet~\cite{zhuo2020effective} & MRNet~\cite{benny2021scale} & \model (Ours) \\ 
    \noalign{\smallskip}
    \hline
    \noalign{\smallskip}
    Acc & 72.5 & 78.3 & 91.4 & 54.5$^\dag$ & 96.6 & \textbf{97.4}   \\
    \noalign{\smallskip}
    \hline
\end{tabular}%
\\
\rule{0in}{1.2em}$^\dag$\scriptsize indicates our evaluation of the baseline in the absence of published results.\\
\end{tabular}
}
\label{tab:raven_fair}
\end{center}
\hfill
\vspace{-2.5em}
\end{table}
\setlength{\tabcolsep}{1.4pt}

\subsection{Performance Analysis} \label{sec:results}

\autoref{tab:Raven_IRaven} summarizes the performance of our model and other baselines on the test set of RAVEN and I-RAVEN datasets. We report the scores from~\cite{zhuo2021unsupervised} for I-RAVEN on LEN, COPINet, and DCNet. We also report the performance of humans on the RAVEN dataset~\cite{zhang2019raven}; there is no reported human performance on I-RAVEN. Overall, our \model achieves superior performance among all baselines for I-RAVEN and a strong performance, on average, on RAVEN. Our method performs similar to DCNet for RAVEN with a slight improvement of $0.4 \%$. We notice DCNet has better performance over ours by a margin of $1.4 \% - 3.5 \%$ over ``3 $\times$ 3 Grid", ``L-R", ``U-D", ``O-IC" and ``O-IG" , while we show $13 \%$ improvement for ``2 $\times$ 2 Grid".
For I-RAVEN, the average test accuracy of our model improves from $95 \%$ (SCL)  to $98.13 \%$ and shows consistent improvement over all configurations across all models. 
The most significant gain, spotted in ``3 $\times$ 3 Grid" and ``O-IG" is expected since our method learns to attend to semantic spatio-visual tokens. As a result, we can focus on the smaller scale objects present in these configurations, which is essential since the attributes of these objects define the rules of the RPM problem. In \autoref{tab:Raven_IRaven}, DCNet achieves better accuracy for ``3 $\times$ 3", ``L-R", ``U-D", ``O-IC", and ``O-IG" for RAVEN but significantly lower accuracy for I-RAVEN, \textbf{suggesting DCNet exploits the shortcut in RAVEN}. See our analysis in \autoref{tab:crossRaven} that supports this observation. 

We also observe that our reasoning model shows more significant improvements on I-RAVEN than RAVEN. This is because the two datasets differ in the selection process of the negative choice set. The wrong images differ in only a single attribute from the right panel for RAVEN, while in I-RAVEN, they differ in at least two characteristics. The latter choice set prevents models from deriving the puzzle solution by only considering the available choices. At the same time, this strategy also helps the classification problem (better I-RAVEN scores) since now the choice set images are more distinct. In \autoref{tab:raven_fair}, we also report the test scores on the RAVEN-FAIR dataset against several baselines. Similar to the above, our model achieves the best performance. 

\autoref{tab:pgm} reports performance of \model and other models trained on the neutral configuration in the PGM dataset. Our model improves by $2.3 \%$ over the best baseline model (SCL). PGM dataset is 20 times larger than RAVEN, and the applied rule can be present either row-wise or column-wise. Additionally, PGM  contains ``line" as an object type, increasing the complexity compared to RAVEN datasets. Even under these additional constraints, \model is able to improve RPM solving by mimicking the reasoning process. 

In \autoref{tab:vprom} we report the performance on the V-PROM dataset, made up of natural images. The background signal for every image in the dataset can be considered as noise or a distractor. In this highly challenging benchmark, our \model shows a major improvement of over $8\%$ over the Relation Network (RN) reported in~\cite{teney2020v} ($51.2$ reported in~\cite{teney2020v} and $52.83$ for our evaluation of the RN model --since the code is not available--). 
Since the V-PROM dataset is the most challenging one, we define the margin $\Delta$ for each testing sample to better understand the performance of \model:
\begin{equation*}\label{eq:delta}
    \Delta = \text{sim} (rc^{12}, rc^{a^*}) - \max_{a \neq a^*} \text{sim}(rc^{12}, rc^a),
\end{equation*}
where $a^*$ indicates the correct answer among eight choices.  The model is confident and correctly answers the RPM question for $\Delta\gg 0$). When $\Delta \approx 0$, the model is uncertain about its predictions, $\Delta<0$ indicating incorrect and $\Delta>0$ correct uncertain predictions. Finally, the model is confident but incorrectly answers for $\Delta\ll 0$).

\begin{figure}[!ht]
    \begin{center}
        \includegraphics[width=0.8\linewidth]{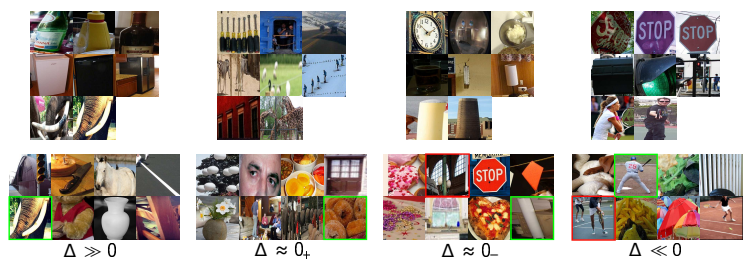}
    \end{center}
    \caption{RPM examples from the V-PROM dataset (trained on \model) for 4-cases of $\Delta$ spanning from correct prediction with strong certainty to incorrect prediction with strong certainty.}
    \label{fig:v-prom_examples}
\end{figure}
In \autoref{fig:v-prom_examples} we present examples from the V-PROM testing set. In the Supplementary we present additional detailed analysis of these results. The first example has a $\Delta>100$, which means that the trained \model is very confident about its prediction.
Next, we move to examples with $\Delta$ very close to zero, either negative or positive. We visualize two such examples in \autoref{fig:v-prom_examples}, in the second (correctly classified RPM) and the third (misclassified) puzzles. The second is a counting problem where the first image in a row has $x$ objects and the next two images in the same row $y$  (i.e., first row $x=7,  y=2$). Some images are distorted and/or blurred after the pre-processing required to use the pre-trained ResNet-101, which makes the recognition and "counting" of objects difficult. The second example is an ``And" rule on object attributes. The first row contains circle objects and the second and third cylinders. The wrongly selected image (depicted in a red bounding box) contains as well cylinder objects. In these two cases, the model is uncertain about which image is solving the RPM.

Lastly, we study examples which \model very confidently misclassifies. Specifically, we picked the worst six instances of the testing set. One of these examples is depicted in the last puzzle of \autoref{fig:v-prom_examples}. All examples belong to the ``And rule" for object attributes like the case $\Delta \gg 0$. In the depicted example, the ``And" rule of the last row refers to "Players"; where in both cases, they are playing "Tennis." The problem is that images five and eight depict "Players" playing "Tennis," resulting in a controversial situation.  Although the model misclassifies the "Player" in the second choice, it is reasonable to choose either the second, fifth, or eighth images as the correct image in the puzzle.

\setlength{\tabcolsep}{5pt}
\begin{table}
\vspace{-1.5em}
\begin{center}
\caption{\hspace*{0mm}Results of cross-dataset evaluation. Top row indicates the training set, next the test set.
}
\resizebox{0.8\columnwidth}{!}{%
\begin{tabular}{ccccccc}
    \hline\noalign{\smallskip}
    Model Training Set   & \multicolumn{2}{c}{RAVEN} & \multicolumn{2}{c}{I-RAVEN} & \multicolumn{2}{c}{RAVEN-F} \\ 
    \noalign{\smallskip}
    Evaluation Dataset & I-RAVEN & RAVEN-F & RAVEN & RAVEN-F & RAVEN & I-RAVEN \\ 
    \noalign{\smallskip}
    \hline
    \noalign{\smallskip}
    SRAN &  72.8 & 78.5 & 57.1 & 71.9 & 54.68 & 60.5  \\
    DCNet & 14.7 & 27.4 & 37.9 & 51.7 & 57.8 & 46.5  \\ 
    \model (Ours) & \textbf{97} & \textbf{97.5} & \textbf{95.1} & \textbf{97.7} & \textbf{94.7} & \textbf{88.3}  \\
    
    \noalign{\smallskip}
    \hline
\end{tabular}%

}
\label{tab:crossRaven}
\end{center}
\hfill
\vspace{-2.5em}
\end{table}
\setlength{\tabcolsep}{1.4pt}

\autoref{tab:crossRaven}  presents our cross-dataset results between models trained and tested on different RAVEN-based datasets.
Since all three datasets only differ in the manner their distractors in the Choice set were created but are identical in the Context, a model close to the generative process should be able to pick the correct image irrespective of how difficult the distractors are. In the first two columns, we train \model with the RAVEN dataset and measure the trained model performance on I-RAVEN, RAVEN-FAIR (RAVEN-F) testing sets; other combinations in the succeeding columns follow the respective train-test patterns. 
As was expected, when training on RAVEN ($94 \%$), we see an improvement on both I-RAVEN ($97 \%$) and RAVEN-FAIR ($97.5 \%$) test since the latter datasets have more dissimilar choice images, helping the reasoning problem. This increase in performance can be seen for SRAN from $56.1 \%$ to $72.8 \%$ and $00 \%$ respectively. Since DCNet ($93.6 \%$) utilizes the short solution, the accuracy drops to $14.7 \%$ and $27.4 \%$ respectively.
For the same reasons, when training on I-RAVEN ($98.8 \%$), our model shows a drop in RAVEN ($95.1 \%$) performance; for RAVEN-FAIR ($97.7 \%$), the performance remains close to the one on the training dataset. 

\subsection{Ablation Study} \label{sec:ablation}
To gain insights into \model learning behavior, we analyze the impact of the local patch receptive field, proxied in $K^2$, the total number of patches, and the input image size $(W,H)$. For a given image of size $W \times H$ and $K \times K$ patches, $\frac{W}{K} \times \frac{H}{K}$ is the size of each patch. In \autoref{tab:analysis}, we fix the patch size to $\frac{W}{K} \times \frac{H}{K} = 32\times 32$, the backbone (ResNet-18), and reasoning networks, and simultaneously vary $K$ and $(W,H)$. In this way, we effectively change the resolution at which the patches model the spatially-contextualized image structures.  Each column represents \model's accuracy for specific $(W,H)$ and $K^2$, e.g., $64, 2 \times 2$ means image is resized to $64 \times 64$ and $2\times 2=4$ image patches of fixed patch size $32 \times 32$) are used.  The results indicate a correlation between the accuracy and the granularity of the patch receptive field; the higher number of patches, with smaller receptive fields aimed at detecting finer spatio-visual image features, the higher the reasoning accuracy.

\setlength{\tabcolsep}{10pt}
\begin{table}
\vspace{-2.0em}
\begin{center}
\caption{\hspace*{0mm}Analysis of sensitivity to Receptive Field Size.
}
\resizebox{0.98\columnwidth}{!}{
\begin{tabular}{l}
\begin{tabular}{cccccccc}
    \hline\noalign{\smallskip}
    Model  & 32, $1 \times 1$ & 64, $2 \times 2$ & 96, $3 \times 3$ & 128, $4 \times 4$ & 160, $5 \times 5$ & 192, $6 \times 6$ & 224, $7 \times 7$ \\ 
    \noalign{\smallskip}
    \hline
    \noalign{\smallskip}
    Acc & 36.0 & 35.6 & 77.0 & 92.4 & 95.9 & 98.2 & 98.1   \\
    \noalign{\smallskip}
    \hline
\end{tabular}%
\\
\end{tabular}
}
\label{tab:analysis}
\end{center}
\hfill
\vspace{-2.5em}
\end{table}
\setlength{\tabcolsep}{1.4pt}

\paragraph{Exploiting shortcut solutions} As shown in SRAN~\cite{hu2020stratified}, any powerful model that learns by combining the extracted features from the choices is capable of exploiting the shortcut solution present in the original RAVEN. In a context-blind setting, a model trained only on images in the RAVEN choice list should predict randomly. However, context-blind $\{$ResNet, CoPINet, DCNet $\}$ models attain $71.9 \%$, $94.2 \%$ and $94.1 \%$ test accuracy respectively. We train and report the accuracy for context-blind DCNet and reported the scores in ~\cite{hu2020stratified} for ResNet, CoPINet. Similarly, we investigate our \model in a context-blind setting. We remove the reasoning module and use the extracted attended choice vectors from the visual transformer, passed through an MLP, to output an eight-dimensional logit vector. After training for 100 epochs, our model performance remained at $12.2\%$, similar to the random guess of $1/8=12.5\%$, suggesting that our ``Backbone" with the ``Visual Transformer" does not contribute towards finding a shortcut. Thus, our semantic tokenized representation coiled with the reasoning module learns rules from the context to solve the RPM questions.


\paragraph{Does \model learn rules?} The rules in RPMs for the I-RAVEN dataset are applied row-wise. However, these rules can exist in  either rows or columns. We evaluate performance on two different setups to determine if our model can discover the rule embeddings with no prior knowledge of whether the rules were applied row-wise or column-wise. In our first setup, we train \model with the prior knowledge of row-wise rules in RPMs. We train our second model by preparing rule embeddings on both row and column and finally concatenating them to predict the correct answer. Our model performance drop was only $3.8 \%$ from $98.1 \%$ to $94.3 \%$.
In case of SRAN~\cite{hu2020stratified}, the reported drop in performance was $1.2 \%$ for the above setting. 
Overall, this indicates that our model is capable of ignoring the distraction from the column-wise rule application.

\begin{figure}[!ht]
     \centering
     \begin{subfigure}[b]{0.24\textwidth}
         \centering
         \includegraphics[width=\textwidth]{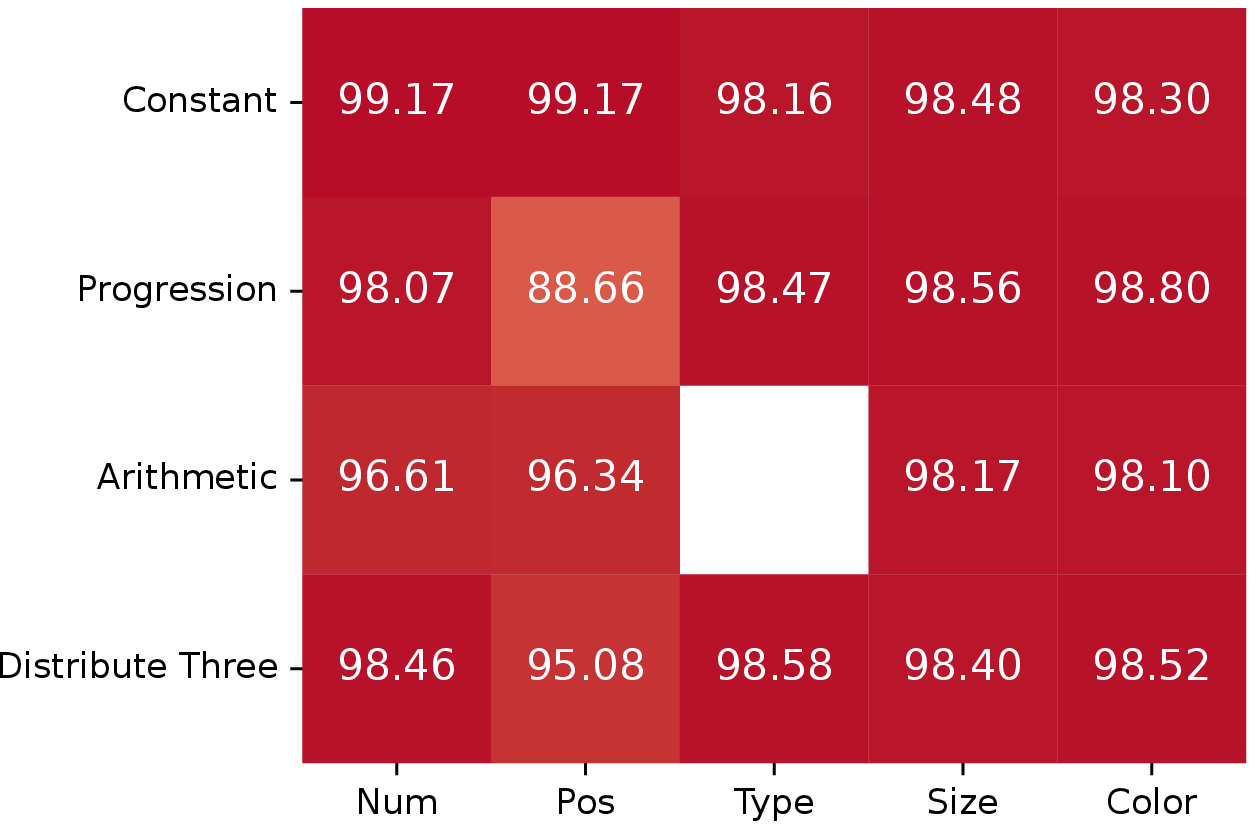}
         \caption{All Configurations}
         \label{fig:all}
     \end{subfigure}
     \hfill
     \begin{subfigure}[b]{0.24\textwidth}
         \centering
         \includegraphics[width=\textwidth]{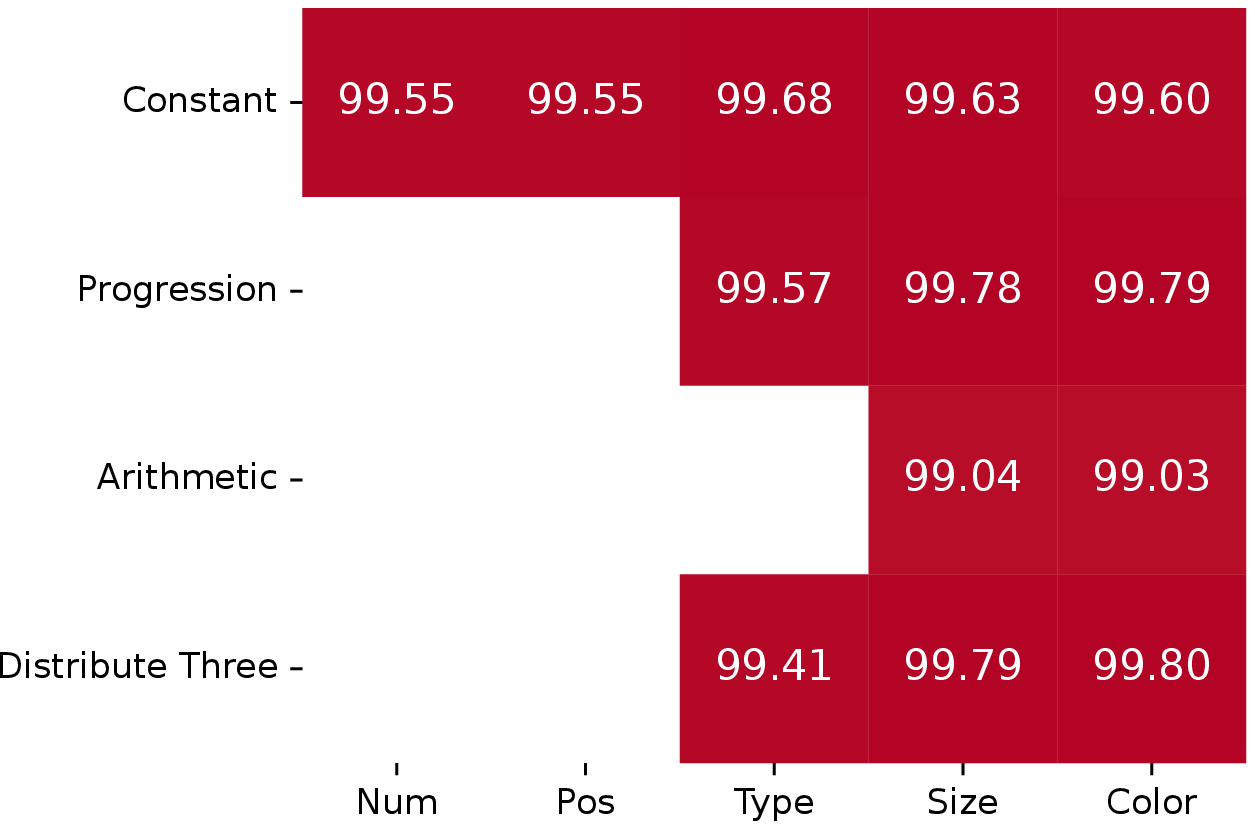}
         \caption{Center Configuration}
         \label{fig:center}
     \end{subfigure}
     \hfill
     \begin{subfigure}[b]{0.24\textwidth}
         \centering
         \includegraphics[width=\textwidth]{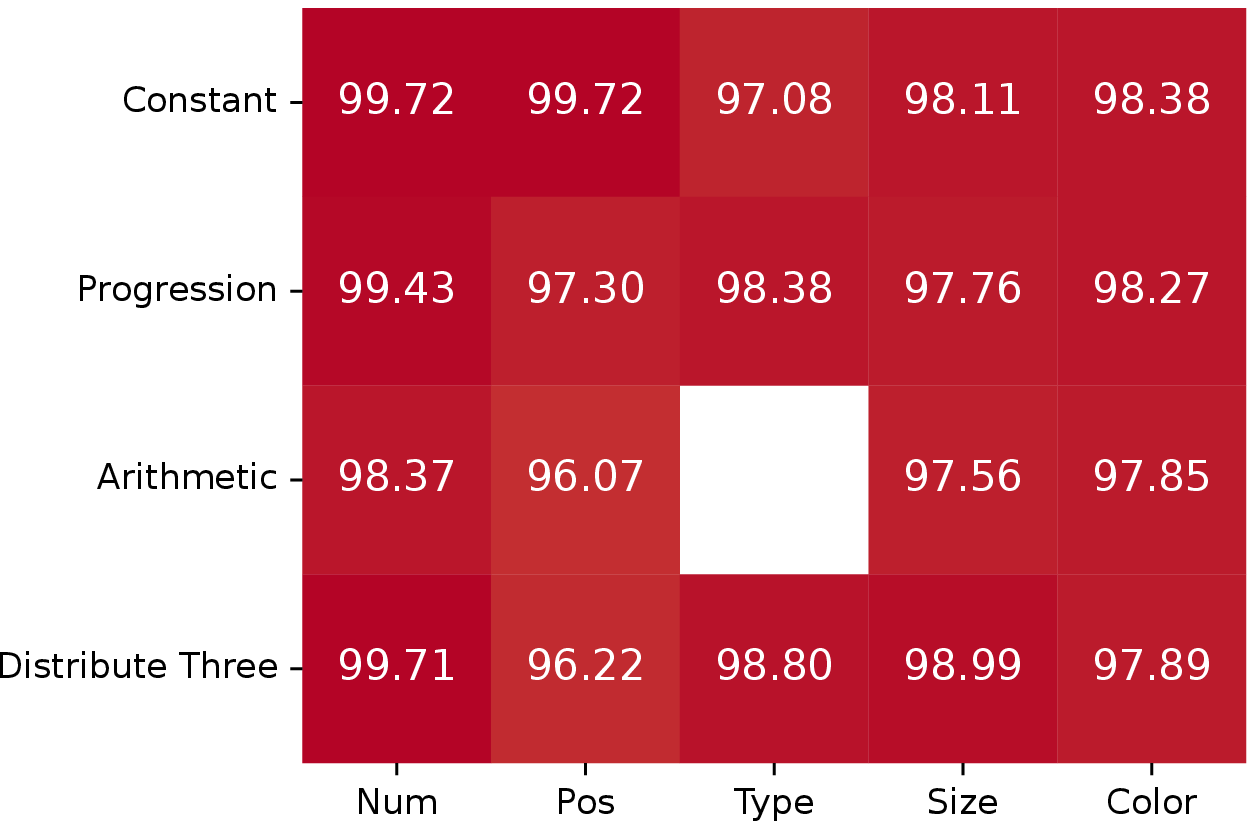}
         \caption{$2 \times 2$ Grid Configuration}
         \label{fig:2x2}
     \end{subfigure}
     \hfill
     \begin{subfigure}[b]{0.24\textwidth}
         \centering
         \includegraphics[width=\textwidth]{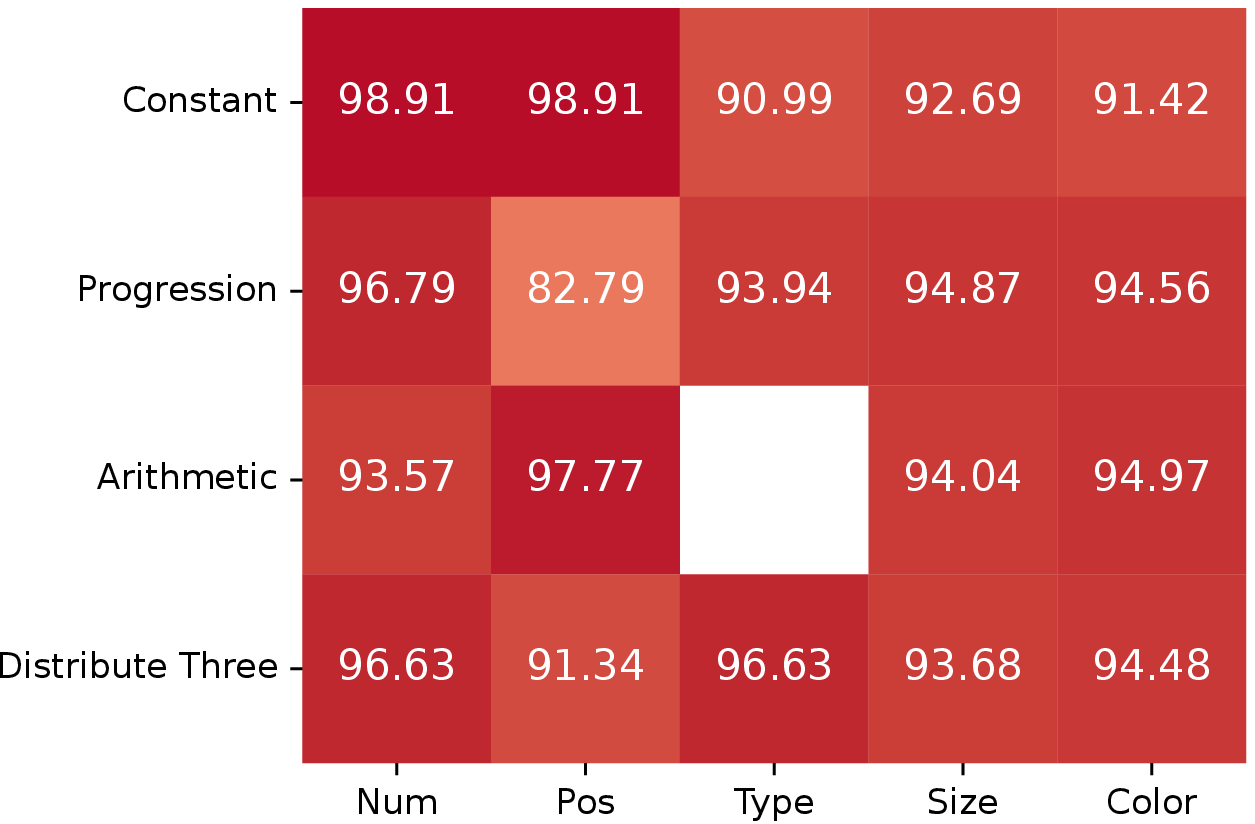}
         \caption{$3 \times 3$ Grid Configuration}
         \label{fig:3x3}
     \end{subfigure}
     \begin{subfigure}[b]{0.24\textwidth}
         \centering
         \includegraphics[width=\textwidth]{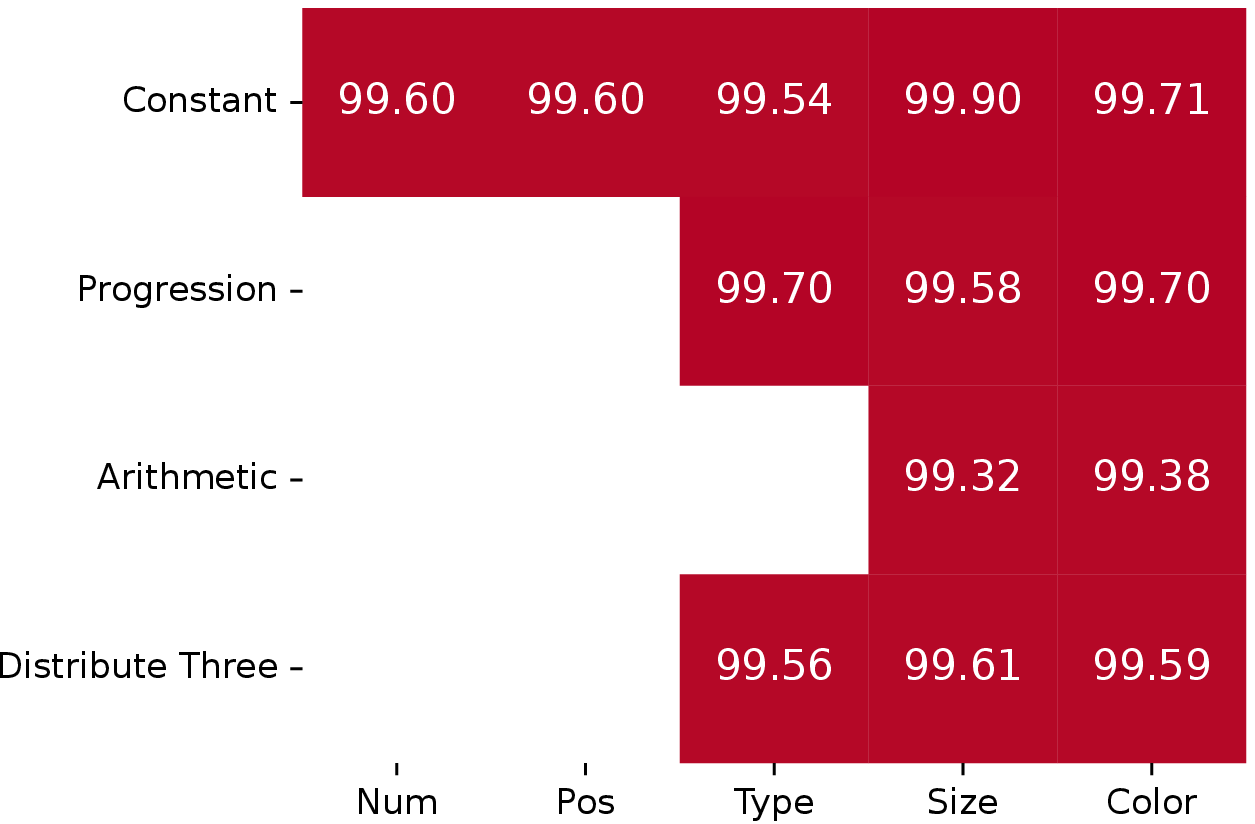}
         \caption{L-R Configuration}
         \label{fig:lr}
     \end{subfigure}
     \hfill
     \begin{subfigure}[b]{0.24\textwidth}
         \centering
         \includegraphics[width=\textwidth]{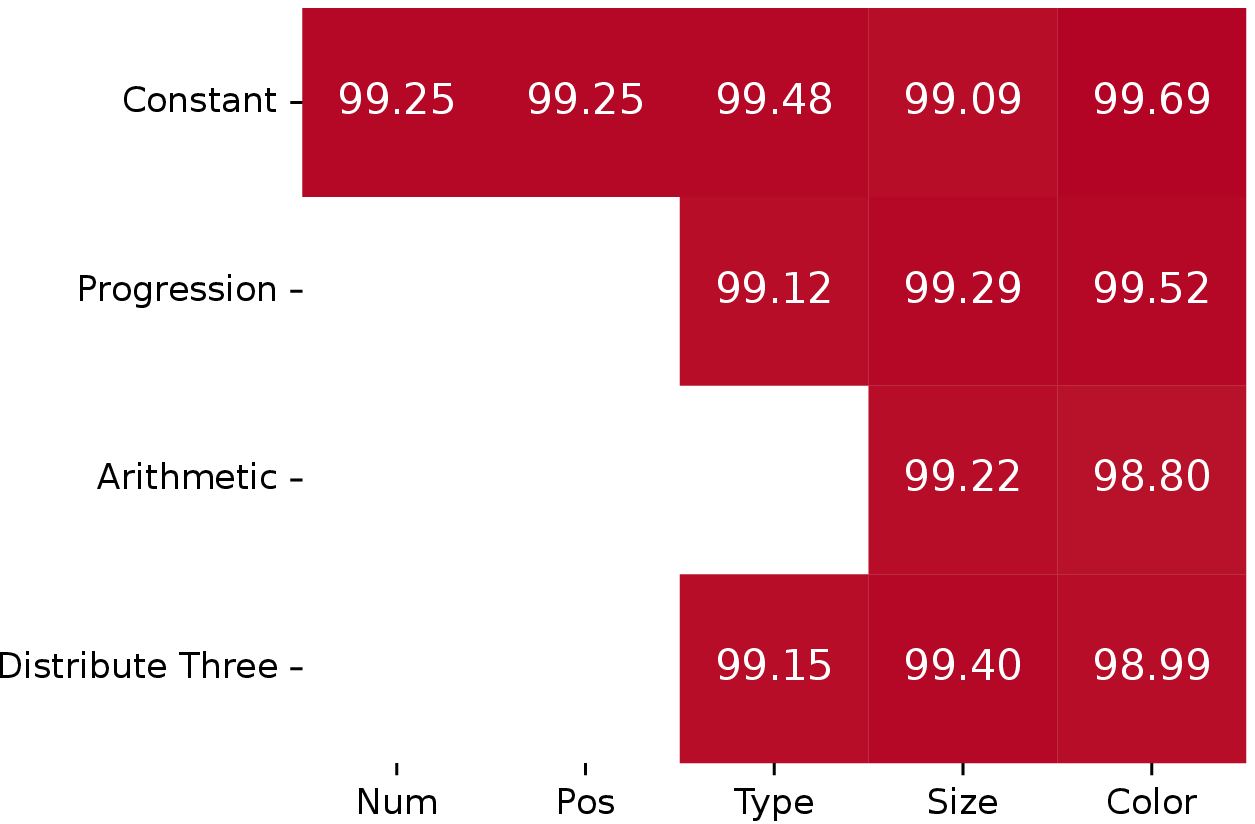}
         \caption{U-D Configuration}
         \label{fig:ud}
     \end{subfigure}
     \hfill
     \begin{subfigure}[b]{0.24\textwidth}
         \centering
         \includegraphics[width=\textwidth]{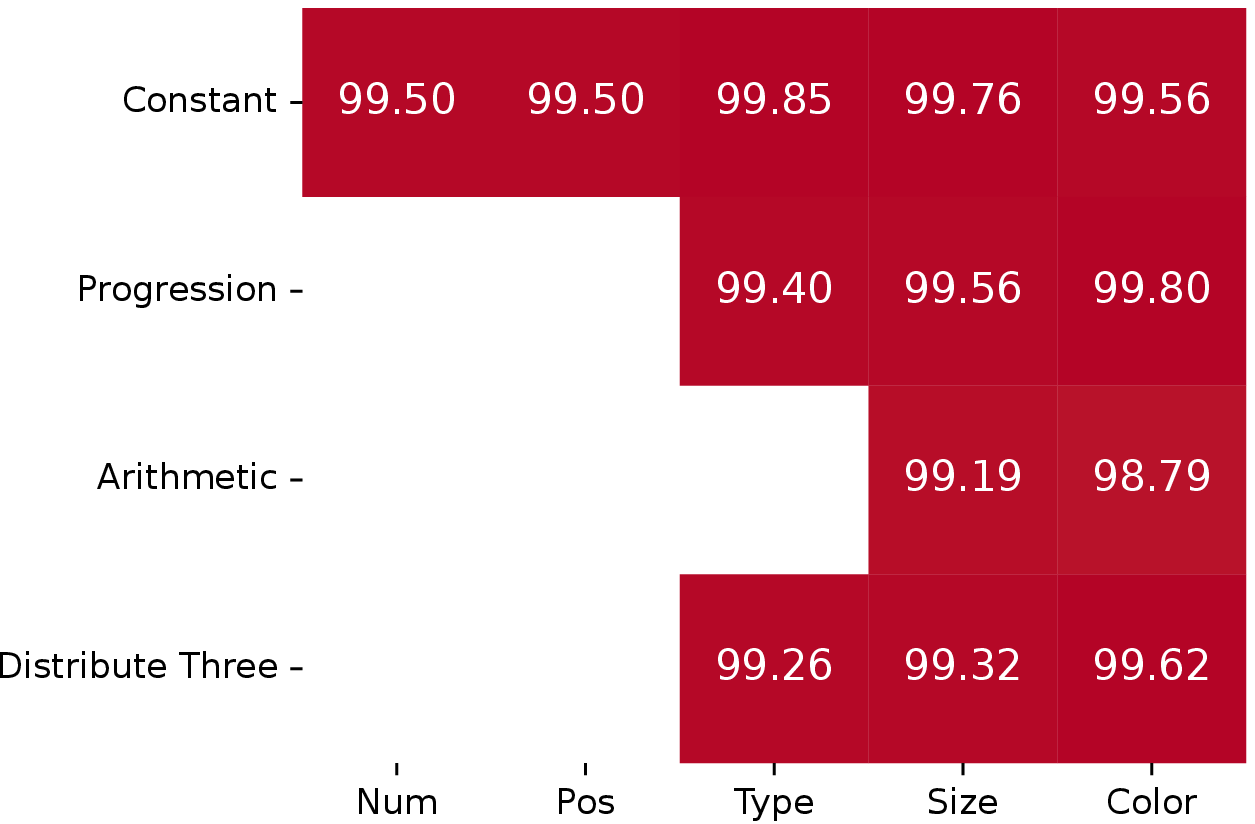}
         \caption{O-IC Configuration}
         \label{fig:oic}
     \end{subfigure}
     \hfill
     \begin{subfigure}[b]{0.24\textwidth}
         \centering
         \includegraphics[width=\textwidth]{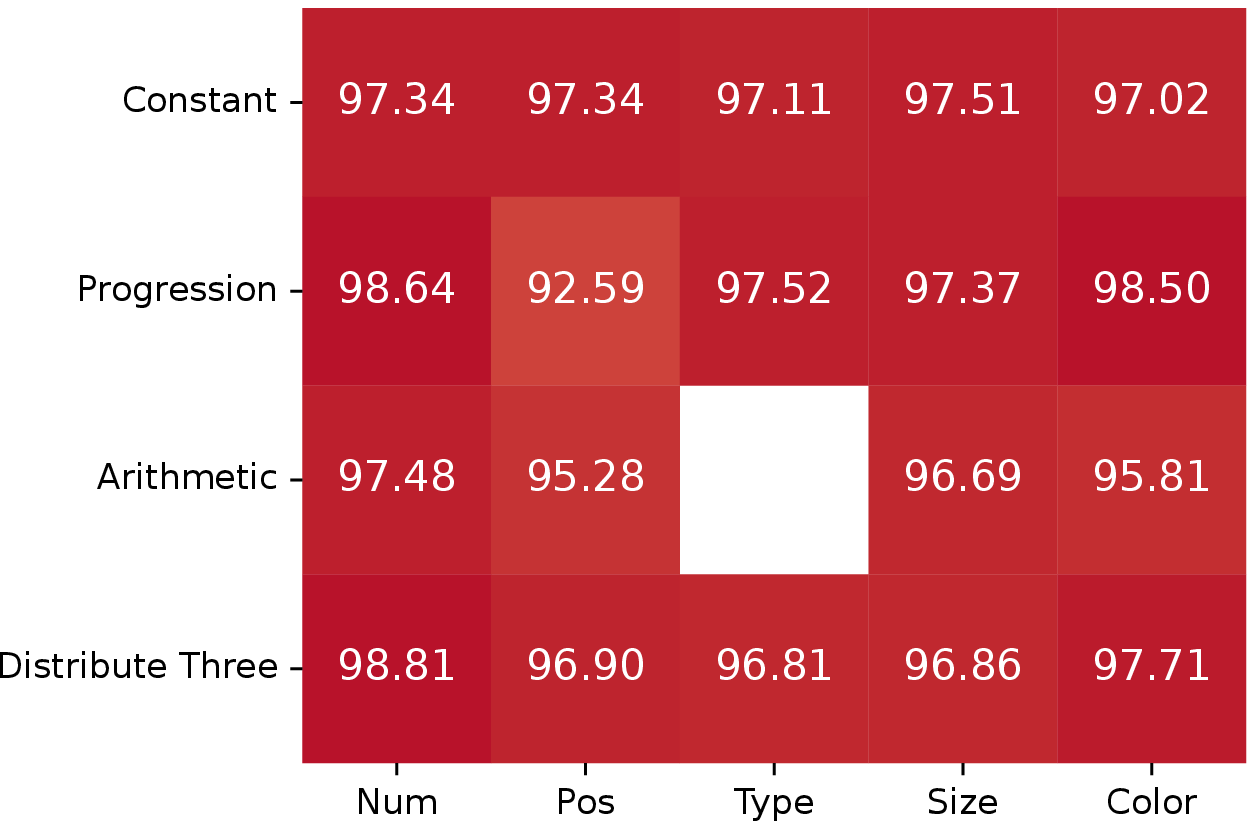}
         \caption{O-IG Configuration}
         \label{fig:oig}
     \end{subfigure}
    \caption{Testing set classification accuracy for the \model trained on all configurations based on each I-RAVEN rule (constant, progression, arithmetic and distribute three) and used attribute (number, position, type, size, color of the objects) . In \autoref{fig:all} we present the classification accuracy for all configurations, in \autoref{fig:center} for center single configuration, in \autoref{fig:2x2} for the $2 \times 2$ Grid, in \autoref{fig:3x3} for the $3 \times 3$ Grid, in \autoref{fig:lr} for the left right, in \autoref{fig:ud} for the up down, in \ref{fig:oic} for the out single, in single center and finally in \autoref{fig:oig} for the out single, in $2\times 2$ Grid.}
    \label{fig:iraven_heatmaps}
\end{figure}

In \autoref{fig:iraven_heatmaps}, we present the I-RAVEN performance on the test set for \model when trained on the I-RAVEN dataset. The eight different heatmap images correspond to the setting with all configurations (\autoref{fig:all}), and individual configurations from ``Center Single"(\autoref{fig:center}) to ``Out-In-Grid" (\autoref{fig:oig}).
The row dimension in each heatmap corresponds to the RPM rules used in the puzzles (Constant, Progression, Arithmetic, and Distribute Three). In I-RAVEN, each rule is associated with an attribute. Therefore, in the columns, we identify the characteristics of different objects, such as their ``Number", ``Position," ``Type," ``Size," and ``Color". The Blank cells in the heatmap e.g., ``(Arithmetic, Type)", indicate non existence for that (rule,attribute) pair combination in the dataset. 

The most challenging combination of (rule,attribute) is a progression with the position. In this setup, the different objects progressively change position on the $2 \times 2$ and $3 \times 3$ grids. The models fail to track this change well. To further understand this drop in performance, we performed extended experiments, investigating the difference in attributes between the correct image and the predicted (misclassified) image (more details in the Appendix). We notice that for the $2\times 2$ grid configuration, the predicted differences are only in one attribute, primarily the ``Position." This means the distractor objects are of the same color, size, type, and number as the correct one but have different positions inside the grid. The second group of misclassified examples differs in the ``Type" of the present objects. Therefore, we can conclude the model finds it challenging to track the proper position of the entities and their type for some examples. In the $3\times 3$ grid, again, ``Position" is the main differentiating attribute, but the ``Size" attribute follows it; this makes sense since, in the $3\times 3$ grid, the objects have a petite size resulting in greater sensitivity to distinguish the different scales. Similar behavior is observed in the O-IG configuration. 
\section{Conclusions}
In this paper, we introduced \model, a model that takes into account the visually-critical spatial context present in image-based RPMs.  By partitioning an image into patches and learning relational reasoning over these local windows, 
our \model fosters learning of Principal rule and attribute representations in RPMs. The model recognizes the Principal Shared rule, comparing it to choices via a simple similarity metric, thus avoiding the possibility of finding a shortcut solution. \model shows robustness to injection of triplets that disobey the RPM formation patterns, e.g., when trained with both choices of row- and column-wise triplets on the RPMs with uniquely, but unknown, row-based rules.  We are the first to provide extensive experiments results on all three RAVEN-based datasets (RAVEN, I-RAVEN, RAVEN-FAIR), PGM, and the challenging natural image-based V-PROM, which suggests that \model outperforms all baselines by a significant margin except for RAVEN where we match their accuracy. 


%
%
%
\bibliographystyle{splncs04}
\bibliography{egbib}

\newpage
\appendix
\section{Appendix}

\subsection{Datasets}
\textbf{RAVEN, I-RAVEN, RAVEN-FAIR} datasets consist of 70K puzzle instances, equally divided into seven distinct configurations: Center, 2x2Grid, 3x3Grid, Left-Right (L-R), Up-Down (U-D), Out-InCenter (O-IC), Out-In Grid (O-IG). An illustration and detailed explanation for each configuration is provided in the Supplement. Each configuration is randomly split into the train, validation, and testing set of ratios 6K:2K:2K. RAVEN dataset has been observed to be biased towards the design of the choice list~\cite{hu2020stratified}. It is shown empirically to discover a shortcut solution to predict the correct choice by only learning from the choice list without the presence of context $\mathcal{M}$. Hence, I-RAVEN and RAVEN-FAIR were proposed to address this by generating the distractor differently. We compare our results against several state-of-the-art models: LSTM~\cite{hochreiter1997long}, WReN~\cite{barrett2018measuring}, ResNet+DRT~\cite{zhang2019raven}, LEN~\cite{zheng2019abstract},
CoPINet~\cite{zhang2019learning}, SRAN~\cite{hu2020stratified}, DCNet~\cite{zhuo2020effective}, and SCL~\cite{wu2020scattering}.

\textbf{PGM} is a large-scale visual reasoning dataset of 1.42M questions, split into the ratio 1.2M : 0.2M : 0.2M for training, validation, and test, respectively. We report our performance of the neutral regime, consisting of samples from the whole set of relationships and visual elements, due to the size of the dataset.
We compare \model against the reported PGM results in WReN, LEN, CoPINet, SRAN, and SCL.

\textbf{V-PROM} is a recently published natural image based RPM dataset derived from the Visual Genome data. It consists of 487,109 examples, where the neutral regime was prepared in the ratio of 2:1 for training and testing. The dataset consists of 84 objects and 38 different human classes of attributes used to create the RPM row rules. The available relations are ``And", ``Or", ``Union", and ``Progression".

\subsection{Model Performance Analysis}

In the Main paper, we present the key evidence in support of the utility of \model over SOTA. In this section, we highlight additional details that reinforce that evidence and point to possible factors behind \model's performance.

Our first aim is to study the robustness of \model and its reasoning predictions. To do this, we define the margin $\Delta$ for each testing sample:
\begin{equation*}\label{eq:supp_delta}
    \Delta = \text{sim} (rc^{12}, rc^{a^*}) - \max_{a \neq a^*} \text{sim}(rc^{12}, rc^a),
\end{equation*}
where $a^*$ indicates the correct answer among the eight choices.  The model is confident and correctly answers the RPM question for a higher positive $\Delta$ value ($\gg 0$). When $\Delta \approx 0$, the model is uncertain about its predictions, $\Delta<0$ indicating incorrect and $\Delta>0$ correct uncertain predictions. Finally, the model is confident but incorrectly answers the RPM question for large negative $\Delta$ values ($\ll 0$).

\begin{figure}[!ht]
    \begin{center}
        \includegraphics[width=\linewidth]{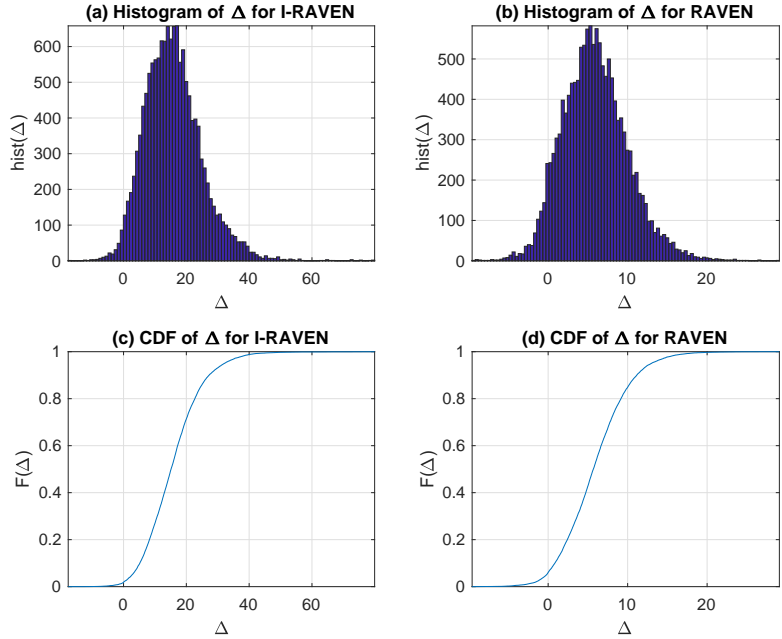}
    \end{center}
    \caption{
    Histogram and CDF plots of the $\Delta$-scores for I-RAVEN and RAVEN trained models. In sub-figures (a) and (b), we can see the histograms of $\Delta$ for I-RAVEN, and RAVEN, respectively. And in (c), (d) the accordingly empirical CDFs.
    }
    \label{fig:delta_iraven_raven}
\end{figure}

\begin{figure}[!ht]
    \begin{center}
        \includegraphics[width=\linewidth]{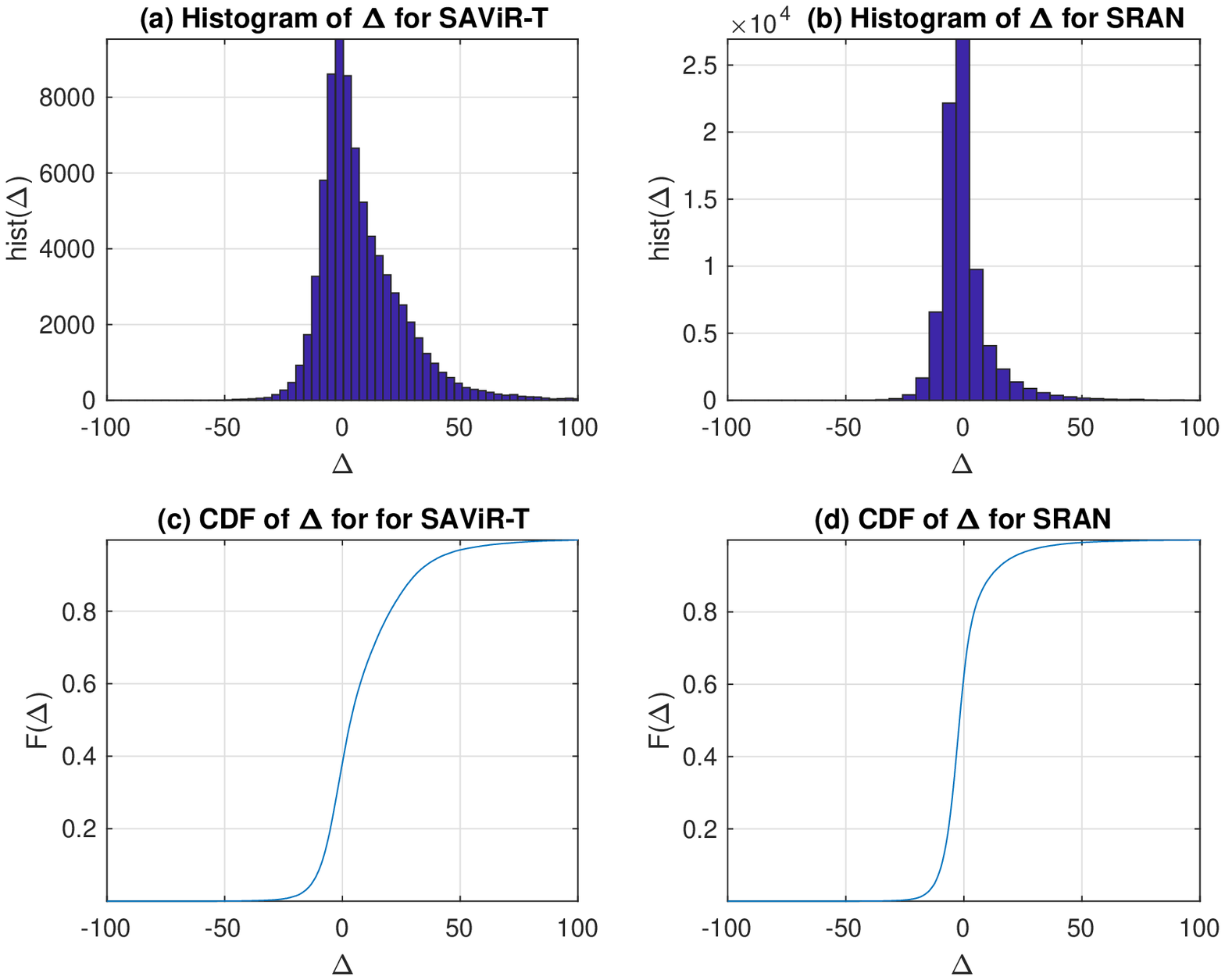}
    \end{center}
    \caption{Histogram and CDF plots of the $\Delta$-scores for the V-PROM dataset. In sub-figures (a) and (b), we can see the histograms of $\Delta$ for \model, and SRAN, respectively. And in (c), (d) the accordingly empirical CDFs.
    }
    \label{fig:delta_vprom}
\end{figure}

\begin{figure}[!ht]
    \begin{center}
        \includegraphics[width=1.0\linewidth]{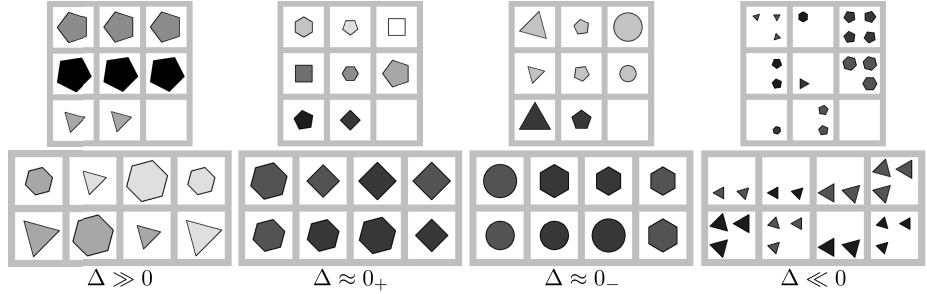}
    \end{center}
    \caption{Sample RPM examples from I-RAVEN dataset (trained on the \model) for four cases of $\Delta$ spanning from correct prediction with strong certainty to incorrect prediction with strong certainty.}
    \label{fig:iraven_delta}
\end{figure}

In \autoref{fig:delta_iraven_raven},\autoref{fig:delta_vprom} 
we present the histogram and empirical CDF of the computed $\Delta$-scores and in \autoref{fig:iraven_delta} we present sample RPM examples for the cases described above. In the case of $\Delta \approx 0$, the predicted image is very similar to the correct image. In the second column, the model elucidated the rule ``decreasing progression'' on object color, ``distribute three'' on object shape (hexagon needed for the last row), and size. On the other hand, in the third column the underlying rule, attribute set in RPM question is $\{$ (constant, number/position), (progression, type), (arithmetic, size), (constant, color)$\}$  the model fails to distinguish between the correct object size even though it had ascertained on the other three pairs.
Finally, the model fails, with high confidence, to predict the correct image for the last column, where the underlying rule, attribute set in RPM question is $\{$ (arithmetic, number), (distribute three, type), (progression, size), (distribute three, color)$\}$, and the model picks up on all rule,attribute pair except for the arithmetic addition on number.

\begin{figure}[!ht]
    \begin{center}
        \includegraphics[width=\linewidth]{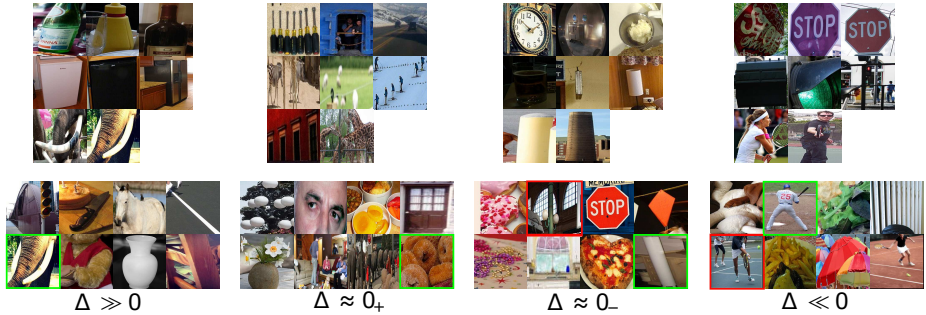}
    \end{center}
    \caption{Sample RPM examples from the V-PROM dataset (trained on the \model) for four cases of $\Delta$ spanning from correct prediction with strong certainty to incorrect prediction with strong certainty.}
    \label{fig:supp_v-prom_examples}
\end{figure}


In \autoref{fig:supp_v-prom_examples} we present some examples from the V-PROM testing set. The first one has a $\Delta$-score greater than 100, which means we are looking at a case where the trained \model model is very confident about its prediction. Although we present a single example for $\Delta>100$, we found six examples in the testing set satisfying the condition. All of them follow the ``And" rule on objects (non-humans). For example, in the first puzzle, the first row corresponds to ``bottle" object attributes, the second to ``fridge," and the last one to ``tusk" meaning that the fifth image having a ``tusk" is solving the RPM puzzle.

Next, we move to examples with $\Delta$-score very close to zero, either negative or positive; in particular, we found twelve examples with $-0.002 < \Delta < 0.002$, with seven correctly classified the rest misclassified. We visualize two such examples in \autoref{fig:supp_v-prom_examples} in the second (correctly classified RPM) and the third (misclassified) puzzle. The second is a counting problem where the first image has $x$ objects and the next to $y$  (i.e., first row $x=7,  y=2$). And the second one is an ``And" rule on object attributes. The first row contains circles and second and third cylinders. The wrongly selected image (depicted in a red bounding box) contains as well cylinder objects. So in these two cases, the model is uncertain about which image is solving the RPM.

Lastly, we have examples which very confidently \model misclassifies. Specifically, we picked the worst six illustrations of the testing set. One of these examples is depicted in the last puzzle of \autoref{fig:supp_v-prom_examples}. All examples belong to the ``And rule" for object attributes like the case $\Delta \gg 0$. In the depicted example, the ``And" rule of the last row refers to "Players"; where in both cases, they are playing "Tennis." The problem is that images five and eight depict "Players" playing "Tennis," resulting in a controversial situation. So although the model misclassifies the "Player" in the second choice, it is reasonable to choose either the second, fifth, or eighth images as the correct image in the puzzle.

\vspace{-0.8em}
\subsection{Sensitivity Analysis}
We additionally train our \model to determine the impact of other components, such as different backbone architecture, size of the backbone network, feature map size, and image size, on the model performance.  We measure reasoning performance against the aforementioned factors while keeping the reasoning model fixed. In \autoref{tab:impact}, we compare against three popular image classification models, AlexNet~\cite{krizhevsky2014one}, VGGNet~\cite{simonyan2014very}, ResNet-18~\cite{he2016deep}. For each architecture, we investigate the impact of the input image resolution using the models denoted as $\{\}-\text{Mini}$, which process scaled-down $64\times 64$ images. We state the output feature map size for all the models. 
Across mini-models, we can notice the lowest performance achieved by AlexNet-Mini ($83.1\%$) with $K^2=16$ and $D=256$, the next highest is for ResNet-18-Mini ($89\%$) with $K^2=25$ and $D=256$ and the best accuracy is achieved by VGGNet-Mini ($91\%$) with $K^2=16$ and $D=512$. For the other three cases, the performance of AlexNet and VGGNet remains nearly the same $94.0\%$ while ResNet-18 scores $98.1\%$. 

\setlength{\tabcolsep}{12.0pt}
\begin{table}
\vspace{-2em}
\centering
\caption{Ablation on the role of backbone model size,  feature map size, backbone architecture, and image size on I-RAVEN.}\label{tab:impact}
\resizebox{0.9\linewidth}{!}{%
\begin{tabular}{lcccc}\\
\hline\noalign{\smallskip}
Model & \begin{tabular}[c]{@{}l@{}} Image \\ Size\end{tabular} & \begin{tabular}[c]{@{}l@{}}Feature Map \\($k \times k$)\end{tabular} & \begin{tabular}[c]{@{}l@{}}\# Output \\ Channels\end{tabular} & Acc \\
\hline\noalign{\smallskip}
AlexNet-Mini & 64 & $4 \times 4$ & 256 & 83.1\\  
AlexNet & 224 & $6 \times 6$ & 256 & 94.4\\ 
VGGNet-Mini & 64 & $4 \times 4$ & 512 & 91\\
VGGNet & 224 & $7 \times 7$ & 512 & 93.9\\
ResNet-18-Mini & 64 & $5 \times 5$ & 256 & 89\\
ResNet-18  & 224 & $7 \times 7$ & 512 & 98.1\\ 
\hline\noalign{\smallskip}
\end{tabular}%
}
\hfill
\vspace{-1.5em}
\end{table}
\setlength{\tabcolsep}{1.4pt}

\subsection{Rule-Attribute Analysis for I-RAVEN}

In \autoref{tab:corvspred} we give the exact number for the ablation study on the failure cases of the testing set for the Progression-Position rule in the I-RAVEN dataset.

\setlength{\tabcolsep}{4.0pt}
\begin{table}[h!]
\caption{Ablation on the failure cases of the testing set for the Progression-Position rule in the I-RAVEN dataset. The first two rows show the total number of Progression-Position rule examples in the Testing set and the number of wrongly predicted RPMs. Next, we count the number of different attributes between the correct choice and the predicted for the misclassified. This is followed by measuring the absolute difference of the exact attributes in which the two images differ. And lastly, since "Color" and "Size" take integer values, we compute the average value difference between the correct and predicted choices.}\label{tab:corvspred}
\centering
\begin{tabular}{lccc}
\hline
\multicolumn{1}{c}{Metrics} & \multicolumn{1}{c}{$2\times 2$ Grid} & \multicolumn{1}{c}{$3 \times 3$ Grid} & \multicolumn{1}{c}{O-IG} \\ \hline
Testing Examples                & 148                              & 308                              & 135                                    \\
Misclassified Examples                  & 17                               & 71                               & 19                                     \\\hline
One Attribute Difference        & 16                               & 62                               & 19                                     \\
Two Attributes Difference      & 1                                & 8                                & 0                                      \\
Three Attributes Difference     & 0                                & 1                                & 0                                      \\\hline
"Number" Attribute Difference  & 0                                & 6                                & 1                                      \\
"Position" Attribute Difference & 9                                & 49                               & 5                                      \\
"Color" Attribute Difference    & 1                                & 8                                & 4                                      \\
"Size" Attribute Difference     & 3                                & 13                               & 5                                      \\
"Type" Attribute Difference     & 5                                & 5                                & 4                                      \\\hline
Average "Color" value difference     & 1                                & 11.08                            & 2                                      \\
Average "Size" value difference       & 1.3                              & 6.05                             & 1.7                                   \\ \hline
\end{tabular}
\end{table}
\vspace{-1.0em}

\subsection{Rule-Attribute Analysis for RAVEN}
\begin{figure}[!ht]
     \centering
     \begin{subfigure}[b]{0.24\textwidth}
         \centering
         \includegraphics[width=\textwidth]{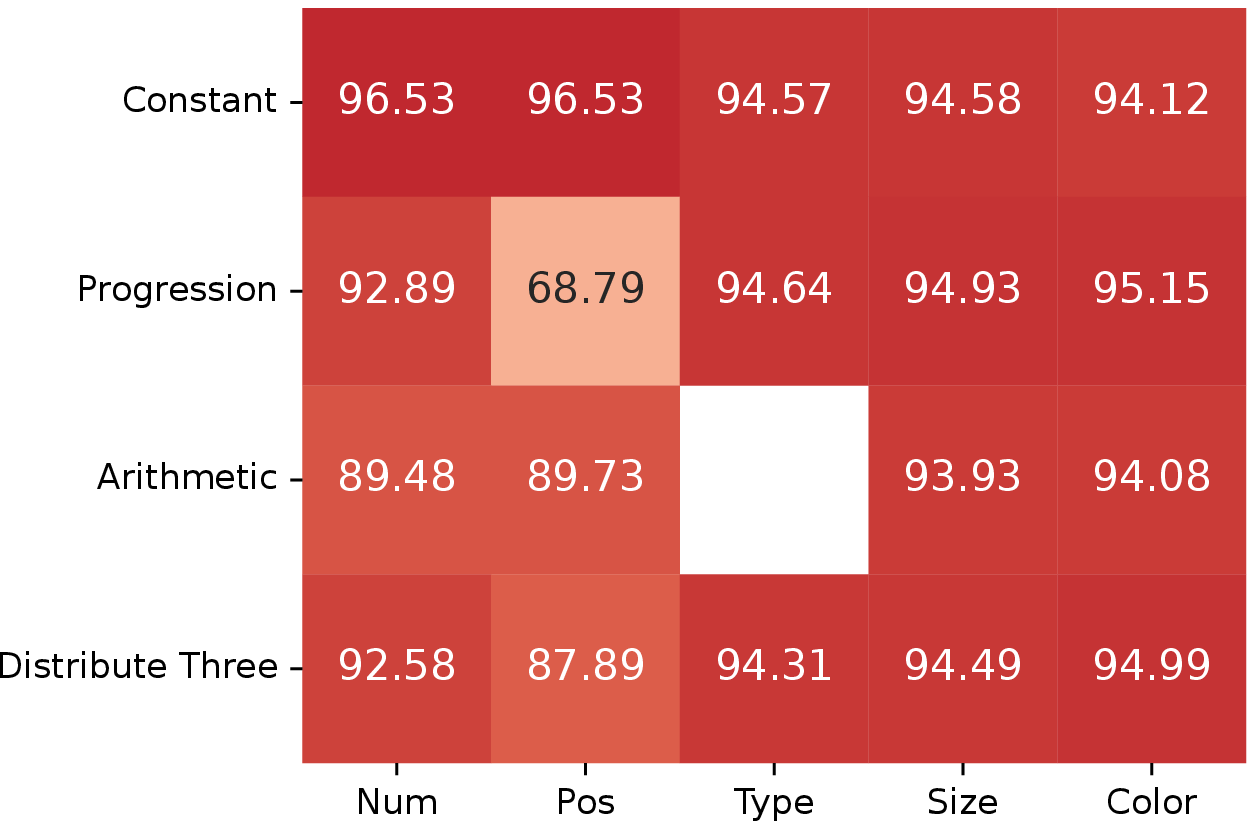}
         \caption{All Configurations}
         \label{fig:raven_all}
     \end{subfigure}
     \hfill
     \begin{subfigure}[b]{0.24\textwidth}
         \centering
         \includegraphics[width=\textwidth]{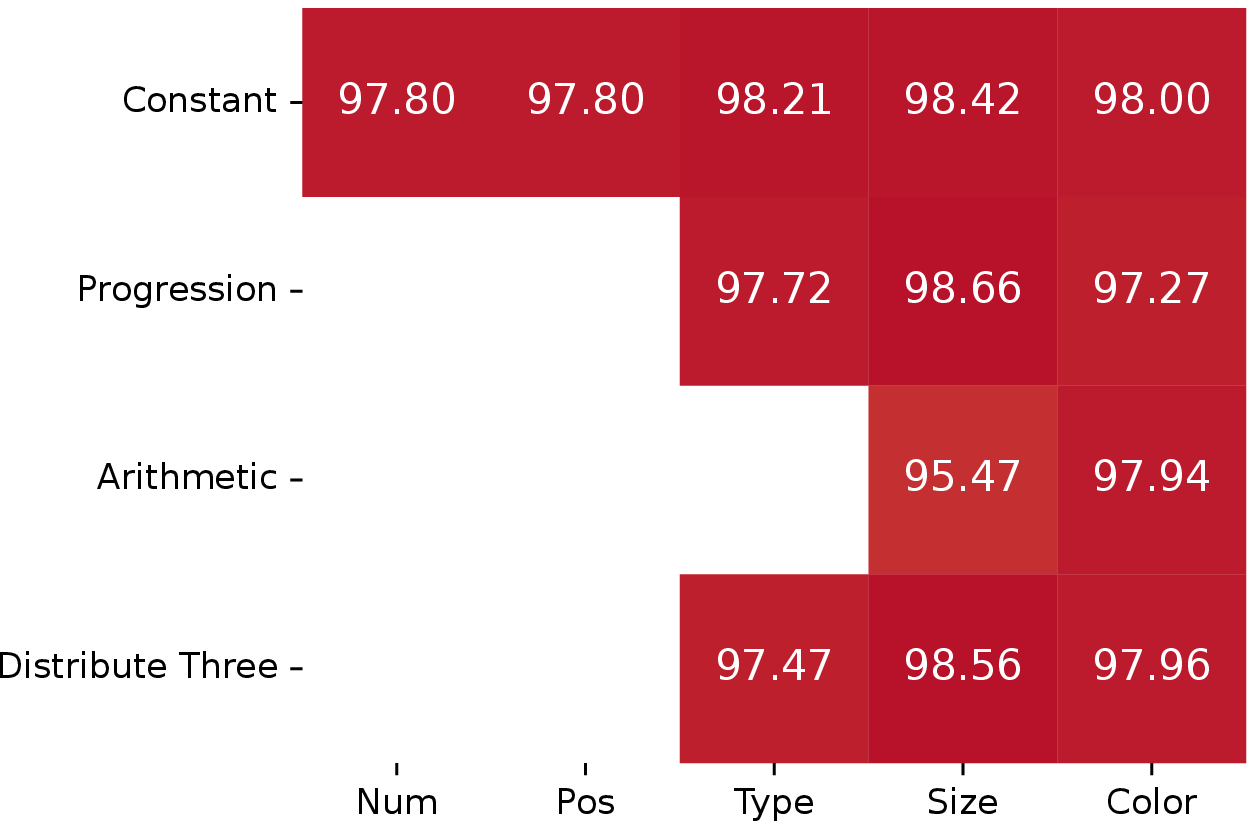}
         \caption{Center Configuration}
         \label{fig:raven_center}
     \end{subfigure}
     \hfill
     \begin{subfigure}[b]{0.24\textwidth}
         \centering
         \includegraphics[width=\textwidth]{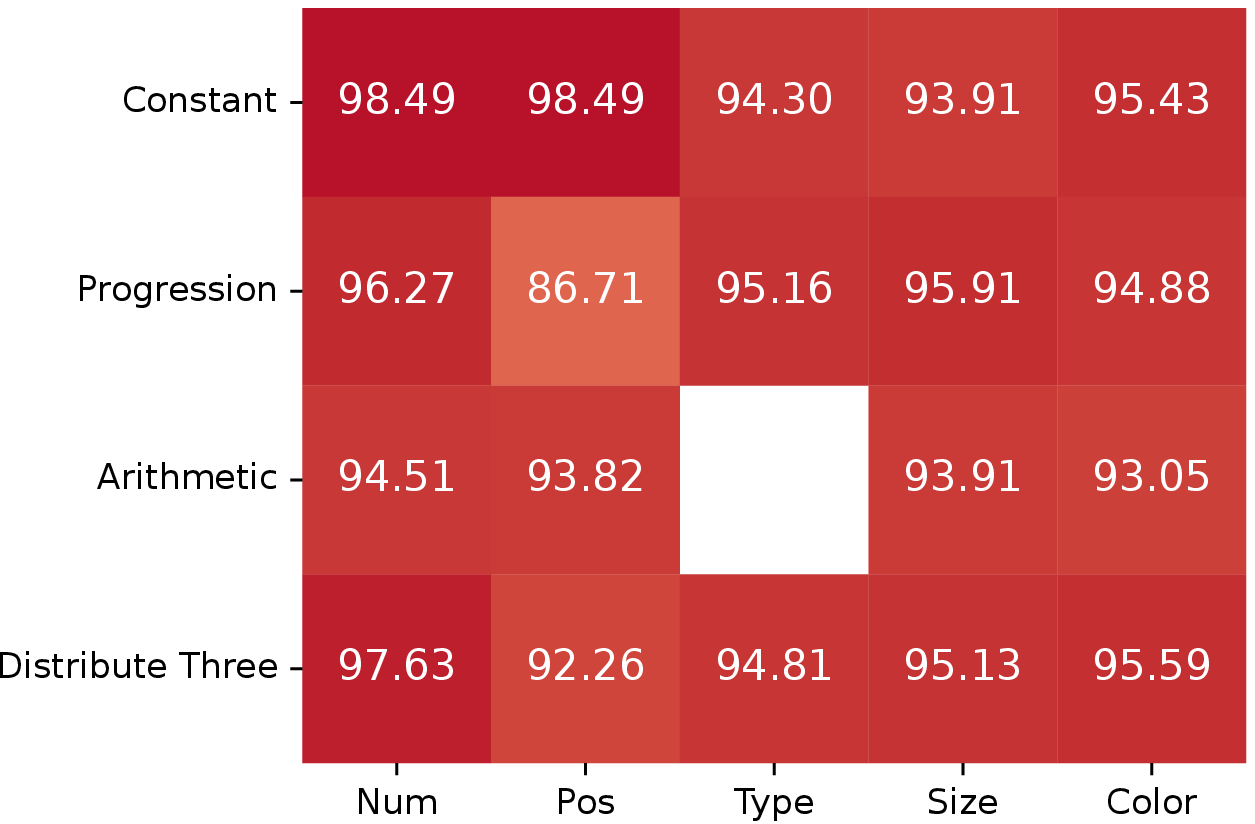}
         \caption{$2 \times 2$ Grid Configuration}
         \label{fig:raven_2x2}
     \end{subfigure}
     \hfill
     \begin{subfigure}[b]{0.24\textwidth}
         \centering
         \includegraphics[width=\textwidth]{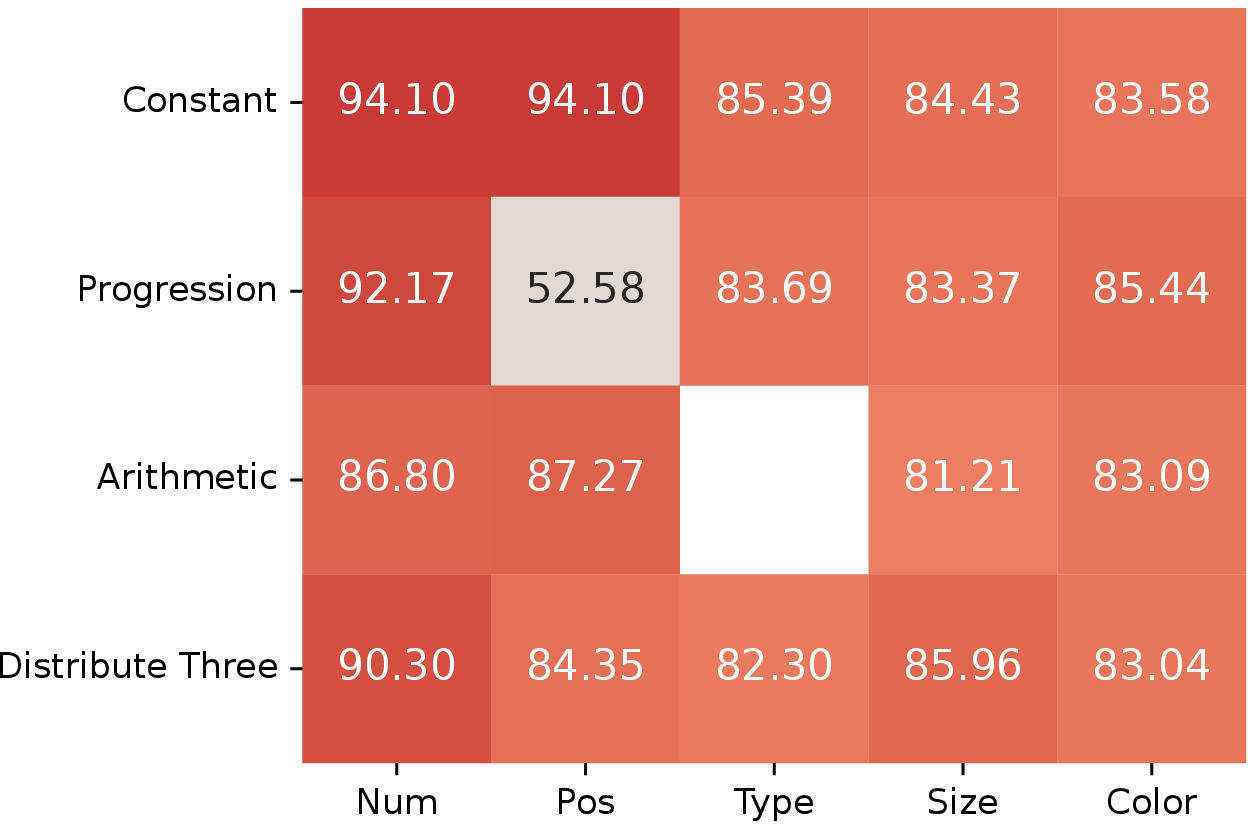}
         \caption{$3 \times 3$ Grid Configuration}
         \label{fig:raven_3x3}
     \end{subfigure}
     \begin{subfigure}[b]{0.24\textwidth}
         \centering
         \includegraphics[width=\textwidth]{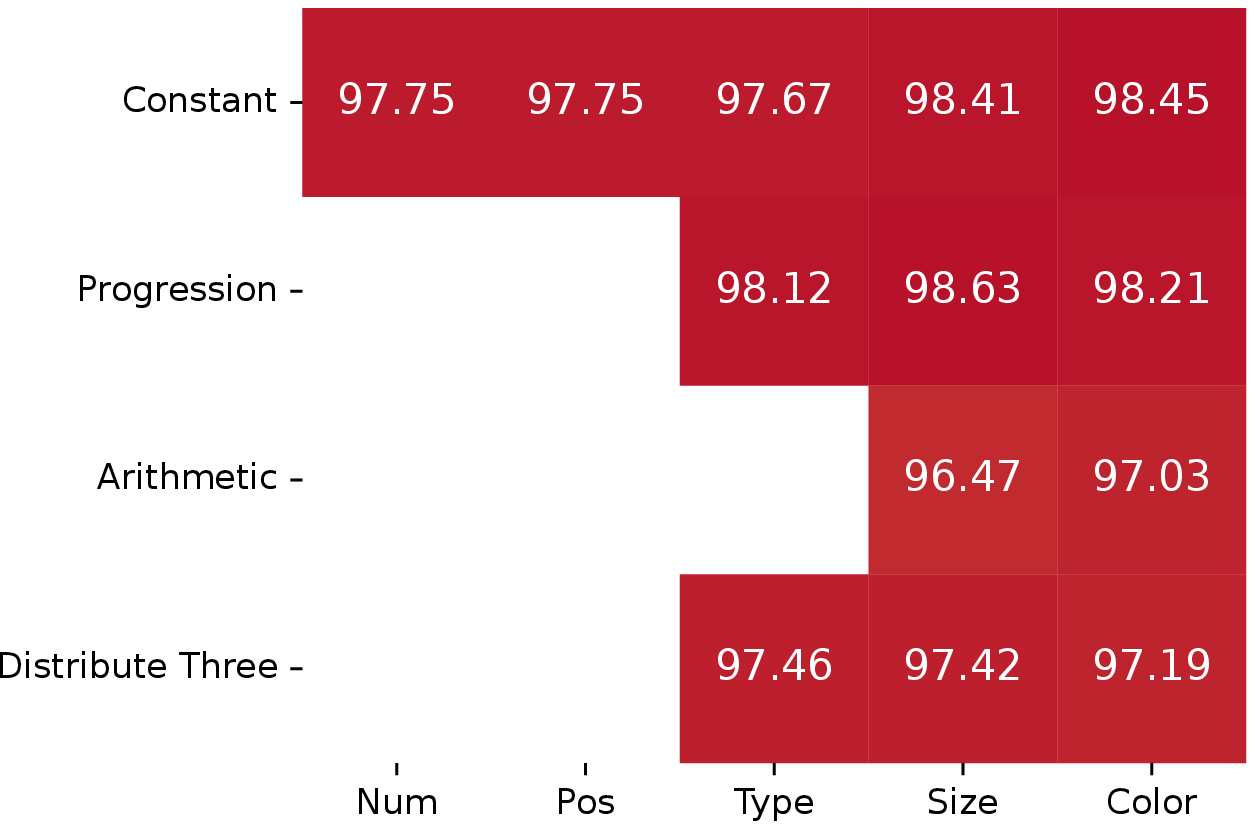}
         \caption{L-R Configuration}
         \label{fig:raven_lr}
     \end{subfigure}
     \hfill
     \begin{subfigure}[b]{0.24\textwidth}
         \centering
         \includegraphics[width=\textwidth]{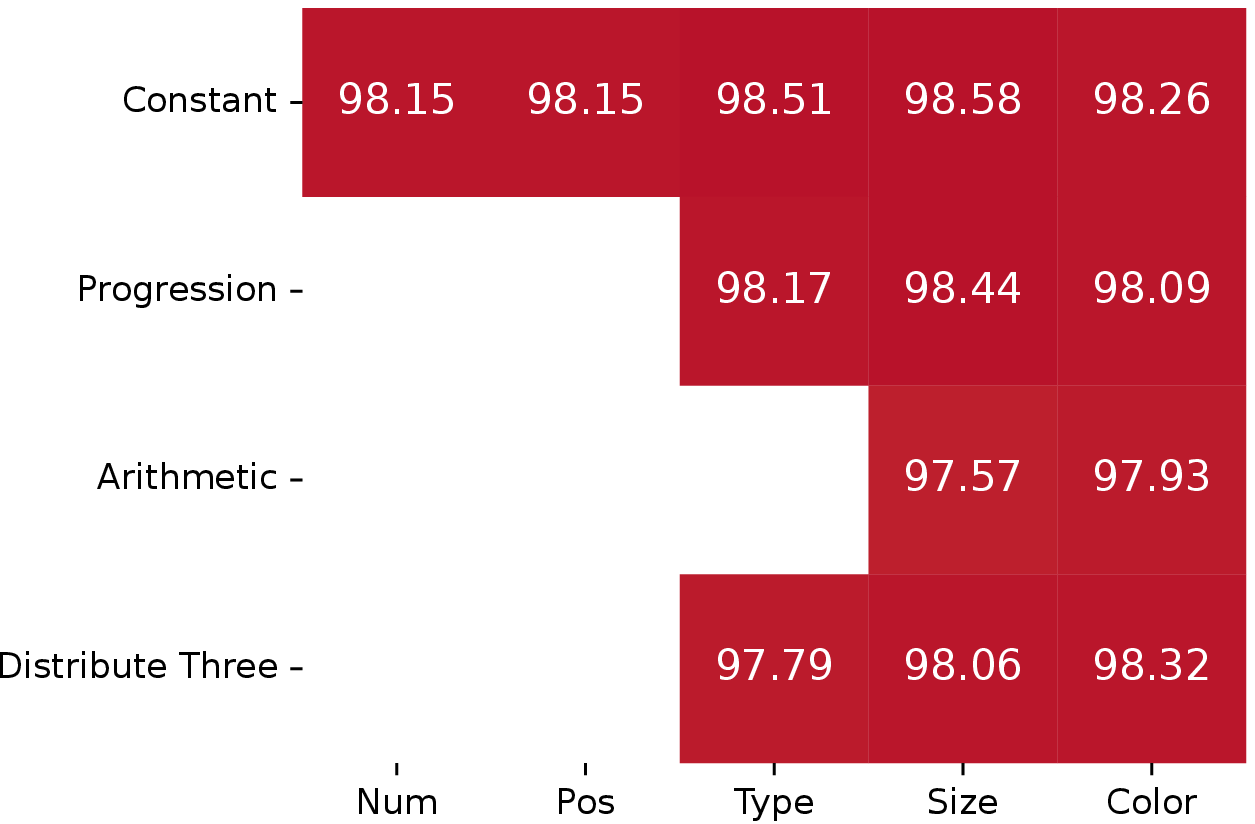}
         \caption{U-D Configuration}
         \label{fig:raven_ud}
     \end{subfigure}
     \hfill
     \begin{subfigure}[b]{0.24\textwidth}
         \centering
         \includegraphics[width=\textwidth]{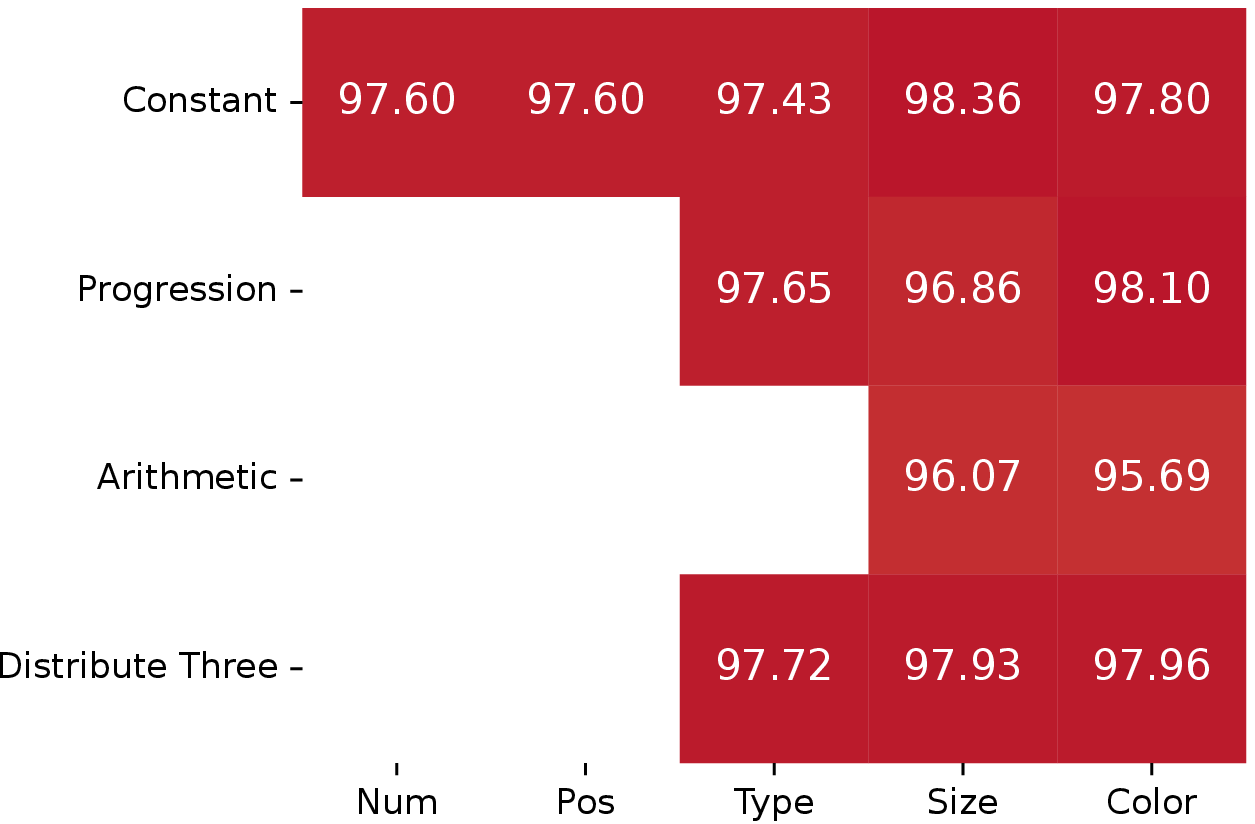}
         \caption{O-IC Configuration}
         \label{fig:raven_oic}
     \end{subfigure}
     \hfill
     \begin{subfigure}[b]{0.24\textwidth}
         \centering
         \includegraphics[width=\textwidth]{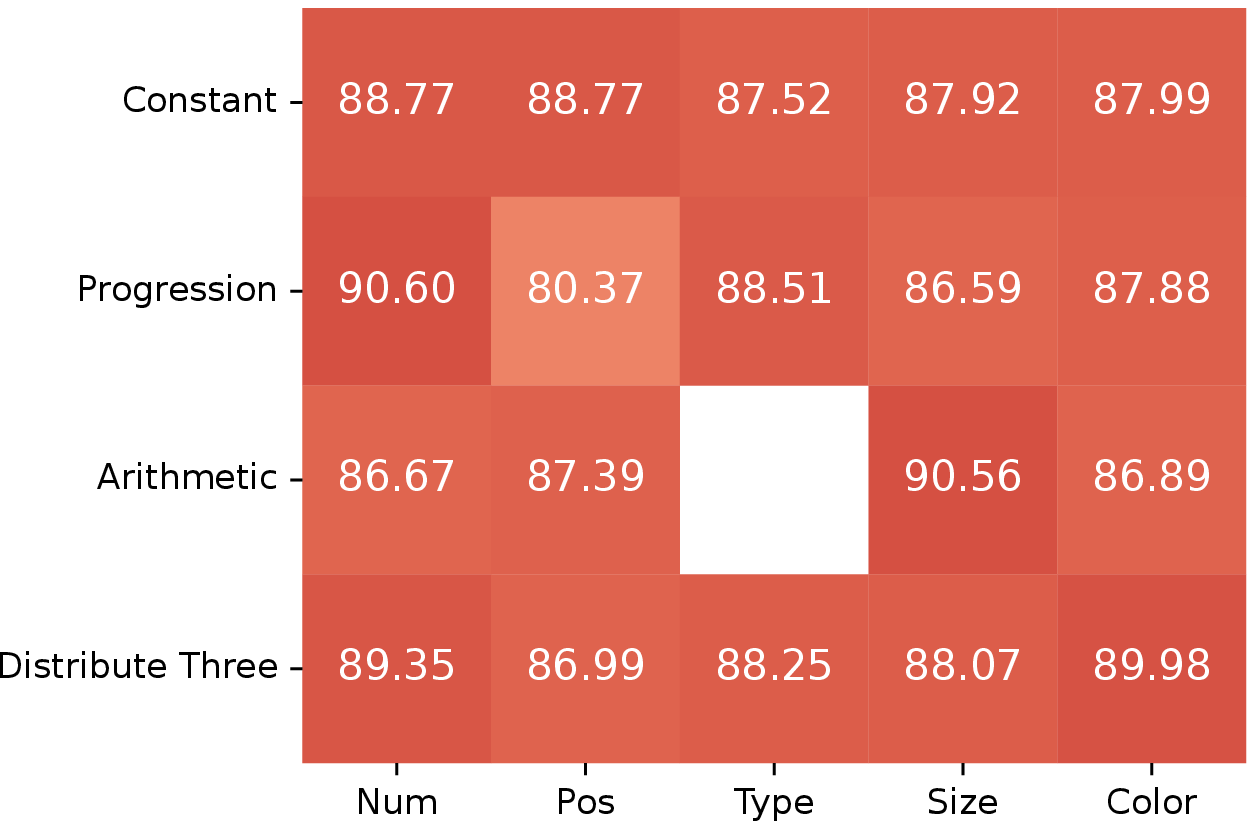}
         \caption{O-IG Configuration}
         \label{fig:raven_oig}
     \end{subfigure}
        \caption{Testing set classification accuracy for the \model trained on all configurations based on each RAVEN rule (constant, progression, arithmetic and distribute three) and used attribute (number, position, type, size, color of the objects) . In \ref{fig:raven_all} we present the classification accuracy for all configurations, in \ref{fig:raven_center} for center single configuration, in \ref{fig:raven_2x2} for the $2 \times 2$ Grid, in \ref{fig:raven_3x3} for the $3 \times 3$ Grid, in \ref{fig:raven_lr} for the left right, in \ref{fig:raven_ud} for the up down, in \ref{fig:raven_oic} for the out single, in single center and finally in \ref{fig:raven_oig} for the out single, in $2\times 2$ Grid.}
        \label{fig:raven_Heatmaps}
\end{figure}

In \autoref{fig:raven_Heatmaps}, we report the performance of \model on RAVEN as  heatmap images for individual configurations and all configuration as a compound. As described for I-RAVEN, we notice our model suffers on RAVEN as well for rule-attribute pair progression, position.

\subsection{V-PROM}

V-PROM is a recently published natural image-based RPM dataset derived from the Visual Genome data. It consists of 487,109 examples where the neutral regime was prepared in the ratio of 2:1 for training and testing. The dataset consists of 84 objects and 38 different human classes of attributes used to create the RPM row rules. The available relations are ``And", ``Or", ``Union", and ``Progression".

As the authors proposed in \cite{teney2020v} instead of using the raw images $224\times 224\times 3$, we use the features extracted from the PyTorch pretrained ResNet-101 before the last average pooling layer; this means that we use for each image a $2048\times 7 \times 7$ representation, which translates to 49 tokens in \model. To further reduce the complexity of our model, we use an MLP layer to derive $512\times7\times 7$ feature vectors; we must mention that this MLP becomes part of \model where we learn its parameters during training.

For the baseline models Relation Network (RN), DCNet, SRAN, we use as input the feature vectors of the PyTorch pretrained ResNet-101 extracted after the last averaging pooling layer (hence a $2048$ feature vector per image). As described in \cite{teney2020v}  the vectors are further normalized by their L2-norm and augmented by the one-hot RPM image encoding. The RPM image encoding refers to the image index in the puzzle (0-15). Consequently, the input of each baseline model is a $2064$ feature vector.

\end{document}